%% file: Arxiv_version.tex
\documentclass[11pt,twoside]{article}

\usepackage{xcolor}

\usepackage{fullpage}

\usepackage{epsf}
\usepackage{fancyhdr}
\usepackage{graphics}
\usepackage{graphicx} 
\usepackage{float} 
\usepackage{subfigure} 
\usepackage{psfrag}
\usepackage{comment}

\usepackage[linesnumbered,ruled]{algorithm2e}% http://ctan.org/pkg/algorithm2e
\DontPrintSemicolon	

\usepackage{color}
\usepackage{amsthm}
\usepackage{amsfonts}
\usepackage{amsmath}
\usepackage{bm}
\usepackage{amssymb,bbm}
\usepackage[numbers]{natbib}
\usepackage{algorithmic}
\usepackage[usestackEOL]{stackengine}
\usepackage{url}
\usepackage[colorlinks=True,linkcolor=magenta,citecolor=blue,urlcolor=blue,pagebackref=true,backref=true]
{hyperref}
\renewcommand*{\backref}[1]{\ifx#1\relax \else Page #1 \fi}
\renewcommand*{\backrefalt}[4]{%
    \ifcase #1 \footnotesize{(Not cited.)}%
    \or        \footnotesize{(Cited on page~#2.)}%
    \else      \footnotesize{(Cited on pages~#2.)}%
    \fi}

\usepackage{nicefrac}

\usepackage{chngpage}

\usepackage{tabularx}%

\usepackage{enumitem}
\usepackage{booktabs}
\usepackage{pbox}

\usepackage{caption}

\usepackage{mathtools}

\usepackage{fullpage}
\allowdisplaybreaks
\input{final_macros.tex}

\def \bq {{\bm q}}
\def \bk {{\bm k}}

\def \bx {{\bm x}}
\def \by {{\bm y}}

\def \bs {{\bm s}}
\def \bt {{\bm t}}

\def \bv {{\bm v}}
\def \bh {{\bm h}}

\def \bQ {{\mathbf Q}}
\def \bK {{\mathbf K}}
\def \bV {{\mathbf V}}
\def \bW {{\mathbf W}}
\def \bA {{\mathbf A}}
\def \bX {{\mathbf X}}
\def \bH {{\mathbf H}}

\newtheorem{remark}{Remark}
\newtheorem{theorem}{Theorem}
\newtheorem{definition}{Definition}
\newtheorem{lemma}{Lemma}

\def\mQ{{\bm{Q}}}

\def\mX{{\bm{X}}}

\def \RR {{\mathbb{R}}}

\def \bX {{\bm X}}
\def \bx {{\bm x}}
\def \by {{\bm y}}
\def \bv {{\bm v}}
\def \bt {{\bm t}}
\def \bv {\mathbf{v}}

\usepackage{verbatim}

\begin{document}
\begin{center}

{\bf{\LARGE{Transformer with Fourier Integral Attentions}}}
  
\vspace*{.2in}
{{
\begin{tabular}{cccccc}
Tan Nguyen$^{\flat, \star}$ & Minh Pham$^{\flat, \star}$ & Tam Nguyen$^{\dagger}$ & Khai Nguyen$^{\dagger}$ & Stanley J. Osher$^{\flat}$ & Nhat Ho$^{\dagger}$ \\
\end{tabular}
}}

\vspace*{.1in}

\begin{tabular}{c}
University of Texas at Austin$^\dagger$, \\
University of California, Los Angeles$^\flat$,\\
\end{tabular}

%{\large{
%\begin{tabular}{ccccc}
%Tongzheng Ren$^{\star, \diamond, \ddag}$ & Fuheng Cui$^{\star,\flat}$ & Alexia Atsidakou$^{\star,\dagger}$ & Sujay Sanghavi$^{\dagger}$ & Nhat Ho$^{\flat, \ddag}$ \\
%\end{tabular}
%}}

%\vspace*{.1in}

%\begin{tabular}{c}
%Department of Computer Science, University of Texas at Austin$^\diamond$, \\
%Department of Statistics and Data Sciences, University of Texas at Austin$^\flat$ \\
%Department of Electrical and Computer Engineering, University of Texas at Austin$^\dagger$, \\
%\end{tabular}

\today

\vspace*{.2in}

\begin{abstract}

Multi-head attention empowers the recent success of transformers, the state-of-the-art models that have achieved remarkable success in sequence modeling and beyond. These attention mechanisms compute the pairwise dot products between the queries and keys, which results from the use of unnormalized Gaussian kernels with the assumption that the queries follow a mixture of Gaussian distribution. There is no guarantee that this assumption is valid in practice. In response, we first interpret attention in transformers as a nonparametric kernel regression. We then propose the FourierFormer, a new class of transformers in which the dot-product kernels are replaced by the novel generalized Fourier integral kernels. Different from the dot-product kernels, where we need to choose a good covariance matrix to capture the dependency of the features of data, the generalized Fourier integral kernels can automatically capture such dependency and remove the need to tune the covariance matrix. We theoretically prove that our proposed Fourier integral kernels can efficiently approximate any key and query distributions. Compared to the conventional transformers with dot-product attention, FourierFormers attain better accuracy and reduce the redundancy between attention heads. We empirically corroborate the advantages of FourierFormers over the baseline transformers in a variety of practical applications including language modeling and image classification.
\end{abstract}
\end{center}

\let\thefootnote\relax\footnotetext{$^{\star}$ Tan Nguyen and Minh Pham contributed equally to this work. Correspondence to: Nhat Ho (\href{mailto:minhnhat@utexas.edu}{minhnhat@utexas.edu}) and Tan Nguyen (\href{mailto:tanmnguyen89@ucla.edu}{tanmnguyen89@ucla.edu}).}
\section{Introduction}
\label{sec:introduction}
%\textbf{Kernel smoothing perspectives of Transformer:} {\color{blue} We will talk about Transformer as a kernel density estimation and nonparametric regression problem.}

%\section{Momentum Networks}\label{sec:momentum:nets}
Transformers~\cite{vaswani2017attention} are powerful neural networks that have achieved tremendous success in many areas of machine learning~\cite{lin2021survey,tay2020efficient,khan2021transformers} and become the state-of-the-art model on a wide range of applications across different data modalities, from language~\cite{devlin2018bert,al2019character,dai2019transformer,child2019generating,JMLR:v21:20-074,baevski2018adaptive,brown2020language,dehghani2018universal} to images~\cite{dosovitskiy2021an,liu2021swin,touvron2020deit,ramesh2021zero,radford2021learning,fan2021multiscale}, videos~\cite{9710415,liu2021video}, point clouds~\cite{zhao2021point,guo2021pct},  and protein sequence~\cite{rives2021biological,jumper2021highly}. In addition to their excellent performance on supervised learning tasks,  transformers can also effectively transfer the learned knowledge from a pretraining task to new tasks with limited or no supervision~\cite{radford2018improving,radford2019language,devlin2018bert,yang2019xlnet,liu2019roberta}. At the core of transformers is the dot-product self-attention, which mainly accounts for the success of transformer models~\cite{cho-etal-2014-learning,parikh-etal-2016-decomposable,DBLP:journals/corr/LinFSYXZB17}. This dot-product self-attention learn self-alignment between tokens in an input sequence by estimating the relative importance of a given token with respect to all other tokens. It then transform each token into a weighted average of the feature representations of other tokens where the weight is proportional to a importance score between each pair of tokens. The importance scores in self-attention enable a token to attend to other tokens in the sequence, thus capturing the contextual representation~\cite{bahdanau2014neural,vaswani2017attention,kim2017structured}. 

\subsection{Self-Attention}
\label{sec:self-attention-def}
Given an input sequence $\bX:=[\bx_1,\cdots,\bx_N]^\top\in \RR^{N\times D_x}$ of $N$ feature vectors, self-attention computes the output sequence $\bH$ from $\bX$ as follows:

\vspace{0.5 em}
\noindent
{\bf Step 1: Projecting the input sequence into different subspaces.} The input sequence $\mX$ is transformed into the query matrix $\bQ$, the key matrix $\bK$, and the value matrix $\bV$ via three linear transformations
    $$
\bQ=\bX\bW_Q^\top; \bK=\bX\bW_K^\top; \bV=\bX\bW_V^\top,
$$
where $\bW_Q,\bW_K\in \RR^{D\times D_x}$, and $\bW_V\in \RR^{D_v\times D_x}$ are the weight matrices. We denote $\mQ:=[\bq_1,\cdots,\bq_N]^\top, \bK:=[\bk_1,\cdots,\bk_N]^\top$, and $\bV:=[\bv_1,\cdots,\bv_N]^\top$, where the vectors $\bq_i,\bk_i,\bv_i$ for $i=1,\cdots,N$ are the query, key, and value vectors, respectively.
    
\vspace{0.5 em}
\noindent
{\bf Step 2: Computing the output as a weighted average.} The output sequence $\bH:=[\bh_1,\cdots,\bh_N]^\top$ is then given by
%     \begin{equation}\label{eq:attention-mat}
% \bH={\rm softmax}\Big(\frac{{\bQ}{\bK}^\top }{\sqrt{D}}\Big){\bf V} :={\bA}{\bV},
% \end{equation}
\begin{equation}\label{eq:attention-mat}
\bH={\rm softmax}\Big({\bQ}{\bK}^\top/\sqrt{D}\Big){\bf V} :={\bA}{\bV},
\end{equation}
where the softmax function is applied to each row of the matrix $(\bQ\bK^\top)/\sqrt{D}$. For each query vector $\bq_i$, $i=1,\cdots,N$, Eqn.~(\ref{eq:attention-mat}) can be written in the vector form to compute the output vector $\bh_i$ as follows 
% \begin{equation}\label{eq:attention-vec}
% \bh_i=\sum_{j=1}^N{\rm softmax}\Big(\frac{{\bq}_i^\top{\bk}_j}{\sqrt{D}}\Big){\bv}_j:=\sum_{j=1}^N a_{ij}\bv_j.
% \end{equation}
\begin{equation}\label{eq:attention-vec}
\bh_i=\sum_{j=1}^N{\rm softmax}\Big({\bq}_i^\top{\bk}_j/\sqrt{D}\Big){\bv}_j:=\sum_{j=1}^N a_{ij}\bv_j.
\end{equation}
The matrix $\bA \in \RR^{N\times N}$ and its component $a_{ij}$ for $i,\,j=1,\cdots,N$  are the attention matrix and attention scores, respectively. The self-attention computed by equations~(\ref{eq:attention-mat}) and~(\ref{eq:attention-vec}) is called the dot-product attention or softmax attention. In our paper, we refer a transformer that uses this attention as the baseline transformer with the dot-product attention or the dot-product transformer. The structure of the attention matrix $\bA$ after training governs the ability of the self-attention to capture contextual representation for each token. 

\vspace{0.5 em}
\noindent
{\bf Multi-head Attention:}
Each output sequence $\bH$ forms an attention head. Multi-head attention concatenates multiple heads to compute the final output. Let $H$ be the number of heads and $\bW^{O} \in \RR^{HD_v \times HD_v}$ be the projection matrix for the output. The multi-head attention is defined as
\begin{equation}
\textrm{MultiHead}(\{\bQ,\bK,\bV\}_{i=1}^{H})=\textrm{Concat}(\bH_1,\dots, \bH_H)\bW^{O}. \nonumber
\end{equation}

The capacity of the attention mechanism and its ability to learn diverse syntactic and semantic relationships determine the success of transformers~\cite{tenney-etal-2019-bert,vig-belinkov-2019-analyzing,clark-etal-2019-bert,voita-etal-2019-analyzing,hewitt-liang-2019-designing}. However, equations~\eqref{eq:attention-mat} and~\eqref{eq:attention-vec} implies that the dot-product attention assumes the features $(q_{i1}, \dots, q_{iD})$ in $\bq_{i}$, as well as the features $(k_{j1}, \dots, q_{jD})$ in $\bk_{j}$, are independent. Thus, the dot-product attention fail to capture the correlations between these features, limiting its representation capacity and inhibit the performance of transformers on practical tasks where there is no guarantee that independent features can learned from complex data. One solution to capture correlations between features $\bq_{i}$ and $\bk_{j}$ is to introduce covariance matrices into the formulation of the dot-product attention with the cost of significantly increasing of the computational complexity. Also, choosing good covariance matrices is difficult.

% Even though multi-head attention extends single-head attention to capture diverse attention patterns and improve the performance of transformers, it has been shown that transformers for practical tasks including sequence classification and language modeling learn redundant heads~\citep{NEURIPS2019_2c601ad9}. These redundant heads compute similar attention mappings. Having many of them in the model limits the representation capacity of the transformer while wasting parameters, memory and computation, impeding the application of transformers to many important large-scale tasks.

\subsection{Contribution} 
In this paper, we first establish a correspondence between self-attention and nonparametric kernel regression. Under this new perspective of self-attention, we explain the limitation of the dot-product self-attention that it may fail to capture correlations between the features in the query and key vectors. We then leverage the generalized Fourier integral theorems, which can automatically capture these correlations, and derive the generalized Fourier integral estimators for the nonparametric regression problem. Using this new density estimator, we propose the FourierFormer, a novel class of transformers that can capture correlations between features in the query and key vectors of self-attention. In summary, our contribution is three-fold:
\begin{enumerate}
    \item We derive the formula of self-attention from solving a nonparametric kernel regression problem, thus providing a nonparametric regression interpretation to study and further develop self-attention.
    \item We develop the generalized Fourier integral estimators for the nonparametric regression problem and provide theoretical guarantees for these estimator.
    \item We propose the FourierFormer whose attentions use the generalized Fourier integral estimators to capture more efficiently correlations between features in the query and key vectors.
\end{enumerate}
Finally, we empirically show that the FourierFormer attains significantly better accuracy than the baseline transformer with the dot-product attention on a variety of tasks including the WikiText language modeling and ImageNet image classsification. We also demonstrate in our experiments that FourierFormer helps reduce the redundancy between attention heads.

\vspace{0.5 em}
\noindent
\textbf{Organization} We structure this paper as follows: In Section~\ref{subsec:regression-intepretation}, we present the correspondence between self-attention and nonparametric kernel regression. In Section~\ref{sec:Fourierformer}, we discuss the generalized Fourier integral estimators and define the FourierFormer. We validate and empirically analyze the advantages of FourierFormer in Section~\ref{sec:experiments}. We
discuss related works in Section~\ref{sec:related_work}. The paper ends with concluding remarks. Technical proofs and
more experimental details are provided in the Appendix.

\vspace{0.5 em}
\noindent
 \textbf{Notation} For any $N \in \mathbb{N}$, we denote $[N] = \{1, 2, \ldots, N\}$. For any $D \geq 1$, $\mathbb{L}_{1}(\mathbb{R}^{D})$ denotes the space of real-valued functions on $\mathbb{R}^{D}$ that are integrable. For any two sequences $\{a_{N}\}_{N \geq 1}, \{b_{N}\}_{N \geq 1}$, we denote $a_{N} = \mathcal{O}(b_{N})$ to mean that $a_{N} \leq C b_{N}$ for all $N \geq 1$ where $C$ is some universal constant.
\section{A Nonparametric Regression Interpretation of Self-attention}
\label{subsec:regression-intepretation}
In this section, we establish the connection between self-attention and nonparametric kernel regression. In particular, we derive the self-attention in equation~\eqref{eq:attention-vec} as a nonparametric kernel regression in which the key vectors $\bk_j$ and value vectors $\bv_j$ are training inputs and training targets, respectively, while the query vectors $\bq_i$ and the output vectors $\bh_i$ form a set of new inputs and their corresponding targets that need to be estimated, respectively, for $i,\,j=1,\cdots,N$. In general, we can view the training set $\{\bk_j, \bv_j\}$ for $j \in [N]$ to come from the following \emph{nonparametric regression model}:
\begin{align}
    \bv_{j} = f(\bk_{j}) + \varepsilon_{j}, \label{eq:nonparametric_regression_softmax_former}
\end{align}
where $\varepsilon_{1}, \ldots, \varepsilon_{N}$ are independent noises such that $\Exs(\varepsilon_{j}) = 0$. Furthermore, we consider a random design setting where the key vectors $\bk_{1}, \bk_{2}, \ldots, \bk_{N}$ are i.i.d. samples from the distribution that admits $p$ as density function. By an abuse of notation, we also denote $p$ as the joint density where the key and value vectors $(\bv_{1}, \bk_{1}), \ldots, (\bv_{N}, \bk_{N})$ are i.i.d. samples from. Here, $f$ is a true but unknown function and we would like to estimate it. 

\vspace{0.5 em}
\noindent
\textbf{Nadaraya–Watson estimator:} Our approach to estimate the function $f$ is based on Nadaraya–Watson's nonparametric kernel regression approach~\cite{Nadaraya64}. In particular, from the nonparametric regression model~\eqref{eq:nonparametric_regression_softmax_former}, we have $\Exs \brackets{\bv_{j}|\bk_{j}} = f(\bk_{j})$ for all $j \in [N]$. Therefore, it is sufficient to estimate the conditional distribution of the value vectors given the key vectors. Given the density function $p$ of the key vectors and the joint density $p$ of the key and value vectors, for any pair of vectors $(\bv, \bk)$ generate from model~\eqref{eq:nonparametric_regression_softmax_former} we have
\begin{align}
    \Exs \brackets{\bv|\bk} = \int_{\mathbb{R}^{D}} \bv \cdot p(\bv|\bk) d\bv = \int \frac{\bv \cdot p(\bv, \bk)}{p(\bk)} d\bv. \label{eq:conditional_expectation}
\end{align}
The formulation~\eqref{eq:conditional_expectation} of the conditional expectation indicates that as long as we can estimate the joint density function $p(\bv, \bk)$ and the marginal density function $p(\bv)$, we are able to obtain an estimation for the conditional expectation and thus for the function $f$. This approach is widely known as Nadaraya–Watson's nonparametric kernel regression approach. 

\vspace{0.5 em}
\noindent
\textbf{Kernel density estimator:} To estimate $p(\bv, \bk)$ and $p(\bk)$, we employ the kernel density estimation approach~\cite{Rosen1956, Parzen62}. In particular, by using the isotropic Gaussian kernel with bandwidth $\sigma$, we have the following estimators of $p(\bv, \bk)$ and $p(\bk)$:
\begin{align}
    \hat{p}_{\sigma}(\bv, \bk) & = \frac{1}{N} \sum_{j = 1}^{N} \varphi_{\sigma}(\bv - \bv_{j}) \varphi_{\sigma}(\bk - \bk_{j}), \quad \quad \hat{p}_{\sigma}(\bk) = \frac{1}{N} \sum_{j = 1}^{N} \varphi_{\sigma}(\bk - \bk_{j}), \label{eq:Gaussian_density_estimator}
\end{align}
where $\varphi_{\sigma}(.)$ is the isotropic multivariate Gaussian density function with diagonal covariance matrix $\sigma^{2} \bold{I}_{D}$. Given the kernel density estimators~\eqref{eq:Gaussian_density_estimator}, we obtain the following estimation of the function $f$:
\begin{align}
    \widehat{f}_{\sigma}(\bk) &= \int_{\mathbb{R}^{D}} \frac{\bv \cdot \hat{p}_{\sigma}(\bv, \bk)}{\hat{p}_{\sigma}(\bk)} d\bv = \int_{\mathbb{R}^{D}} \frac{\bv \cdot \sum_{j = 1}^{N} \varphi_{\sigma}(\bv - \bv_{j}) \varphi_{\sigma}(\bk - \bk_{j})}{\sum_{j = 1}^{N} \varphi_{\sigma}(\bk - \bk_{j})} d\bv \nonumber \\
    & = \frac{\sum_{j = 1}^{N} \phi_{\sigma}(\bk - \bk_{j}) \int \bv \cdot  \varphi_{\sigma}(\bv - \bv_{j}) d\bv}{\sum_{j = 1}^{N} \varphi_{\sigma}(\bk - \bk_{j})} 
     = \frac{\sum_{j = 1}^{N} v_{j} \varphi_{\sigma}(\bk - \bk_{j})}{\sum_{j = 1}^{N} \varphi_{\sigma}(\bk - \bk_{j})}. \label{eq:Gaussian_nonparametric_regression}
\end{align}
\textbf{Connection between Self-Attention and nonparametric regression:} By plugging the query vectors $\bq_{i}$ into the function $\widehat{f}_{\sigma}$ in equation~\eqref{eq:Gaussian_nonparametric_regression}, we obtain that
\begin{align}
    \widehat{f}_{\sigma}(\bq_{i}) &= \frac{\sum_{j}^{N}\bv_{j}\exp\left(-\|\bq_{i} - \bk_{j}\|^{2}/2\sigma^2\right)}{ \sum_{j}^{N}\exp\left(-\|\bq_{i} - \bk_{j}\|^{2}/2\sigma^2\right)} \nonumber \\
    &= \frac{\sum_{j}^{N}\bv_{j}\exp\left[-\left(\|\bq_{i}\|^{2} + \|\bk_{j}\|^{2}\right)/2\sigma^{2}\right]\exp\left(\bq_{i}\bk_{j}^{\top}/\sigma^{2}\right)}{\sum_{j}^{N}\exp\left[-\left(\|\bq_{i}\|^{2} + \|\bk_{j'}\|^{2}\right)/2\sigma^{2}\right]\exp\left(\bq_{i}\bk_{j}^{\top}/\sigma^{2}\right)}. \label{eqn:attn_reg}
\end{align}
If we further assume that the keys $\bk_{j}$ are normalized, which is usually done in practice to stabilize the training of transformers~\cite{schlag2021linear}, the value of $\widehat{f}_{\sigma}(\bq_{i})$ in equation~\eqref{eq:Gaussian_nonparametric_regression} then becomes
\begin{align}
    \widehat{f}_{\sigma}(\bq_{i}) = \frac{\sum_{j}^{N}\bv_{j}\exp\left(\bq_{i}\bk_{j}^{\top}/\sigma^{2}\right)}{\sum_{j'}^{N}\exp\left(\bq_{i}\bk_{j'}^{\top}/\sigma^{2}\right)} 
                        = \sum_{j=1}^N{\rm softmax}\Big({\bq}_i^\top{\bk}_j/\sigma^{2}\Big){\bv}_j. \label{eqn:attn_reg_final}
\end{align}
When we choose $\sigma^{2} = \sqrt{D}$ where $D$ is the dimension of ${\bq}_i$ and $\bk_{j}$, equation~\eqref{eqn:attn_reg_final} matches equation~\eqref{eq:attention-vec} of self-attention, namely, $\widehat{f}_{\sigma}(\bq_{i}) = \bh_{i}$. Thus, we have shown that self-attention performs nonparametric regression using isotropic Gaussian kernels. 

\begin{remark}
\label{rm:gaussian-vs-dotproduct}
The assumption that $\bk_{j}$ is normalized is to recover the pairwise dot-product attention in transformers. In general, this assumption is not necessary. In fact, the isotropic Gaussian kernel in equation~\eqref{eqn:attn_reg} is more desirable than the dot-product kernel in equation~\eqref{eqn:attn_reg_final} of the pairwise dot-product attention since the former is Lipschitz while the later is not Lipschitz~\cite{kim2021lipschitz}. The Lipschitz constraint helps improve the robustness of the model~\cite{cisse2017parseval,tsuzuku2018lipschitz,anil2019sorting} and stabilize the model training~\cite{miyato2018spectral}.
\end{remark}

\vspace{0.5 em}
\noindent
{\bf Limitation of Self-Attention:} From our nonparametric regression interpretation, self-attention is derived from the use of isotropic Gaussian kernels for kernel density estimation and nonparametric regression estimation, which may fail to capture the complex correlations between $D$ features in ${\bq}_i$ and $\bk_{j}$~\cite{WJones93, Ho21}. Using multivariate Gaussian kernels with dense covariance matrices can help capture such correlations; however, choosing good covariance matrices is challenging and inefficient~\cite{Wand92, Stanis93, Chacon18}. In the following section, we discuss the Fourier integral estimator and its use as a kernel for computing self-attention in order to overcome these limitations.

\section{FourierFormer: Transformer via Generalized Fourier Integral Theorem}\label{sec:Fourierformer}
 In the following, we introduce generalized integral theorems that are able to capture the complex interactions among the features of the queries and keys.  We then apply these theorems to density estimation and nonparametric regression problems. We also establish the convergence rates of these estimators. Given these density estimators, we introduce a novel family of transformers, named \emph{FourierFormer}, that integrates the generalized Fourier integral theorem into the dot-product attention step of the standard transformer. 

\subsection{(Generalized) Fourier Integral Theorems and Their Applications}
\label{subsec:generalized-fourier}
The Fourier integral theorem is a beautiful result in mathematics~\cite{Wiener33, Bochner_1959} and has been recently used in nonparametric mode clustering, deconvolution problem, and generative modeling~\cite{Ho21}. It is a combination of Fourier transform and Fourier inverse transform. In particular, for any function $p \in \mathbb{L}_{1}(\mathbb{R}^{D})$, the \emph{Fourier integral theorem} is given by
\begin{align}
    p(\bk) & = \frac{1}{(2\pi)^{D}} \int_{\mathbb{R}^{D}} \int_{\mathbb{R}^{D}} \cos(\bs^{\top}(\bk - \by)) p(\by) d\by d\bs \nonumber \\ 
    & = \frac{1}{\pi^{D}} \lim_{R \to \infty} \int_{\mathbb{R}^{D}} \prod_{j = 1}^{D} \frac{\sin(R(k_{j} - y_{j}))}{(k_{j} - y_{j})} p(\by) d\by, \label{eq:Fourier_integral_theorem}
\end{align}
where $\bk = (k_{1}, \ldots, k_{D})$ and $\by = (y_{1}, \ldots, y_{D})$. Equation~\eqref{eq:Fourier_integral_theorem} suggests that $$p_{R}(\bk) : = \frac{1}{\pi^{D}} \int_{\mathbb{R}^{D}} \prod_{j = 1}^{D} \frac{\sin(R(y_{j} - k_{j}))}{(y_{j} - k_{j})} p(\by) d\by$$ can be used as an estimator of the function $p$. 

\vspace{0.5 em}
\noindent
\textbf{Benefits of the Fourier integral over Gaussian kernel:} There are two important benefits of the estimator $p_{R}$: (i) it can automatically preserve the correlated structure lying within $p$ even when $p$ is very complex and high dimensional function. It is in stark contrast to the standard kernel estimator built based on multivariate Gaussian kernel where we need to choose good covariance matrix in the multivariate Gaussian kernel to guarantee such estimator to work well. We note that as the standard soft-max Transformer is constructed based on the multivariate Gaussian kernel, the issue of choosing good covariance matrix in dot-product transformer is inevitable; (ii) The product of sinc kernels in the estimator $p_{R}$ does not decay to a point mass when $R \to \infty$. It is in stark difference from the multivariate Gaussian kernel estimator, which converges to a point mass when the covariance matrix goes to 0. It indicates that $p_{R}$ is a non-trivial estimator of the function $p$. Finally, detailed illustrations of these benefits of the Fourier integral over Gaussian kernel in density estimation and nonparametric regression problems, which we have just shown to have connection to the self-attention in transformer, can be found in Section 8 in~\cite{Ho21}.

\vspace{0.5 em}
\noindent
\textbf{Generalized Fourier integral estimator:} Borrowing the above benefits of Fourier integral estimator $p_{R}$, in the paper we would like to consider a generalization of that estimator, named \emph{generalized Fourier integral estimator}, which is given by:
\begin{align}
    p_{R}^{\phi}(\bk) : = \frac{R^{D}}{A^{D}} \int_{\mathbb{R}^{D}} \prod_{j = 1}^{D} \phi \parenth{\frac{\sin(R(y_{j} - k_{j}))}{R(y_{j} - k_{j})}} p(\by) d\by, \label{eq:generalized_Fourier_integral_estimator}
\end{align}
where $A : = \int_{\mathbb{R}} \phi \left(\frac{\sin(z)}{z}\right) dz$ and $\phi: \mathbb{R} \to \mathbb{R}$ is a given function. When $\phi(\bk) = \bk$ for all $\bk \in \mathbb{R}^{D}$, the generalized Fourier integral estimator $p_{R}^{\phi}$ becomes the Fourier integral estimator $p_{R}$. Under appropriate conditions on the function $\phi$ (see Theorem~\ref{theorem:density_estimation_first} in Section~\ref{sec:density_estimation_Fourier} and Theorem~\ref{theorem:density_estimation_second} in Appendix~\ref{sec:additional_theory_Generalized Fourier density estimator}), the estimator $p_{R}^{\phi}$ converges to the true function $p$, namely,
\begin{align}
    p(\bk) = \lim_{R \to \infty} p_{R}^{\phi}(\bk) = \lim_{R \to \infty} \frac{R^{D}}{A^{D}} \int_{\mathbb{R}^{D}} \prod_{j = 1}^{D} \phi \parenth{\frac{\sin(R(y_{j} - k_{j}))}{R(y_{j} - k_{j})}} p(\by) d\by. \label{eq:generalized_Fourier_integral_theorem}
\end{align}
We name the above limit as \emph{generalized Fourier integral theorem}. Furthermore, the estimator $p_{R}^{\phi}$ also inherits similar aforementioned benefits of the Fourier integral estimator $p_{R}$. Therefore, we will use the generalized Fourier integral theorem as a building block for constructing density estimators and nonparametric regression estimators, which are crucial to develop the FourierFormer in Section~\ref{sec:Transformer_generalized_Fourier}.
\subsubsection{Density Estimation via Generalized Fourier Integral Theorems}
\label{sec:density_estimation_Fourier}
 We first apply the generalized Fourier integral theorem to the density estimation problem. To ease the presentation, we assume that $\bk_{1}, \bk_{2}, \ldots, \bk_{N} \in \mathbb{R}^{D}$ are i.i.d. samples from a distribution admitting density function $p$ where $D \geq 1$ is the dimension.  Inspired by the generalized Fourier integral theorem, we obtain the following \emph{generalized Fourier density estimator} $p_{N,R}^{\phi}$ of $p$ as follows:
\begin{align}
    p_{N, R}^{\phi}(\bk) : = \frac{R^{D}}{N A^{D}} \sum_{i = 1}^{N} \prod_{j = 1}^{D} \phi \left(\frac{\sin(R(k_{j} - k_{ij}))}{R(k_{j} - k_{ij})} \right), \label{eq:generalized_Fourier_density_estimator}
\end{align}
where $A = \int_{\mathbb{R}} \phi \left(\frac{\sin(z)}{z}\right) dz$ and $\bk_{i} = (k_{i1}, \ldots, k_{iD})$ for all $i \in [N]$. To quantify the error between the generalized Fourier density estimator $p_{n,R}^{\phi}$ and the true density $p$, we utilize mean integrated squared errors (MISE)~\cite{Wass06}, which is given by:
\begin{align}
    \text{MISE}(p_{N, R}^{\phi}, p) : = \int_{\mathbb{R}^{D}} (p_{N, R}^{\phi}(\bk) - p(\bk))^2 d\bk. \label{eq:MISE}
\end{align}
We start with the following bound on the MISE between $p_{n,R}^{\phi}$ and $p$.
\begin{theorem}
\label{theorem:density_estimation_first}
Assume that $\int_{\mathbb{R}} \phi(\sin(z)/z)z^{j} dz = 0$ for all $j \in [m]$ and $\int_{\mathbb{R}} |\phi(\sin(z)/z)| |z|^{m + 1} dz < \infty$ for some $m \in \mathbb{N}$. Then, there exist universal constants C and C' depending on $d$ and $A$ such that
\begin{align*}
    \text{MISE}(p_{N, R}^{\phi}, p) \leq \frac{C}{R^{m + 1}} + \frac{C' R^{D}}{N}. 
\end{align*}
\end{theorem}
Proof of Theorem~\ref{theorem:density_estimation_first} is in Appendix~\ref{subsec:proof:theorem:density_estimation_first}. A few comments are in order. First, by choosing $R$ to balance the bias and variance in the bound of MISE in Theorem~\ref{theorem:density_estimation_first}, we have the optimal $R$ as $R = \mathcal{O}(N^{1/(D + m + 1)})$. With that choice of $R$, the MISE rate of $p_{N, R}^{\phi}$ is $\mathcal{O}(N^{-(m+1)/(D + m + 1)})$. Second, when $\phi(z) = z^{l}$ for $l \geq 4$ and $z \in \mathbb{R}$, the assumptions in Theorem~\ref{theorem:density_estimation_first} are satisfied when $m = 1$. Under this case, the MISE rate of $p_{N,R}^{\phi}$ is $\mathcal{O}(N^{-2/(D+2)})$. However, these assumptions do not satisfy when $\phi(z) = z^{l}$ and $l \in \{1,2,3\}$, which is due to the limitation of the current proof technique of Theorem~\ref{theorem:density_estimation_first} that is based on Taylor expansion of the estimator $p_{n,R}^{\phi}$.

To address the limitation of the Taylor expansion technique, we utilize the Plancherel theorem in Fourier analysis to establish the MISE rate of $p_{N,R}^{\phi}$ when $\phi(z) = z^{l}$ and $l \in \{1,2,3\}$. The details of the theoretical analyses for such setting are in Appendix~\ref{sec:additional_theory}.
\subsection{FourierFormer: Transformers with Fourier Attentions}
\label{sec:Transformer_generalized_Fourier}
Motivated by the preservation of the correlated structure of the function from the generalized Fourier integral theorem as well as the theoretical guarantees of density estimators, in this section we adapt the nonparametric regression interpretation of self-attention in Section~\ref{subsec:regression-intepretation} and propose the generalized Fourier nonparametric regression estimator in Section~\ref{sec:nonparametric_regression_Fourier}. We also establish the convergence properties of that estimator. Then, based on generalized Fourier nonparametric regression estimator, we develop the Fourier Attention and its corresponding FourierFormer in Section~\ref{sec:fourierformer_definition}.

\subsubsection{Nonparametric Regression via Generalized Fourier Integral Theorem}
\label{sec:nonparametric_regression_Fourier}
We now discuss an application of the generalized Fourier integral theorems to the nonparametric regression setting~\eqref{eq:nonparametric_regression_softmax_former}, namely, we assume that $(\bv_{1}, \bk_{1}), \ldots, (\bv_{N}, \bk_{N})$ are i.i.d. samples from the following nonparametric regression model:
\begin{align*}
    \bv_{j} = f(\bk_{j}) + \varepsilon_{j},
\end{align*}
where $\varepsilon_{1}, \ldots, \varepsilon_{N}$ are independent noises such that $\Exs(\varepsilon_{j}) = 0$ and the key vectors $\bk_{1}, \bk_{2}, \ldots, \bk_{N}$ are i.i.d. samples from $p$. Given the generalized Fourier density estimator~\eqref{eq:generalized_Fourier_density_estimator}, following the argument in Section~\ref{subsec:regression-intepretation}, the Nadaraya–Watson estimator of the function $f$ based on the generalized Fourier density estimator is given by:
%assume that $(Y_{1}, X_{1}), \ldots, (Y_{n}, X_{n}) \in \mathbb{R}^{d + 1}$ are i.i.d. samples from the following nonparametric regression model: $Y_{i} = f(X_{i}) + \epsilon_{i}$ for any $i \in [n]$ where $\epsilon_{1}, \ldots, \epsilon_{n}$ are independent noises such that $\Exs(\epsilon_{i}) = 0$ and $\var(\epsilon_{i}) = \sigma^2$ for some given variance $\sigma > 0$. Here, $f$ is a true but unknown function and we would like to estimate it. Furthermore, we consider a random design setting where $X_{1}, X_{2}, \ldots, X_{n}$ are i.i.d. samples from the distribution that admits $p_{0}$ as density function. 
\begin{align}
    f_{N,R}(\bk) : = \frac{\sum_{i = 1}^{N} \bv_{i} \prod_{j = 1}^{D} \phi \left(\frac{\sin(R(k_{j} - k_{ij}))}{R(k_{j} - k_{ij})} \right)}{\sum_{i = 1}^{N} \prod_{j = 1}^{D} \phi \left(\frac{\sin(R(k_{j} - k_{ij}))}{R(k_{j} - k_{ij})} \right)}. \label{eq:nonparametric_regression_Fourier}
    %: = \frac{a_{N, R}(x)}{p_{N, R}(x)}, 
\end{align}
%where $p_{n, R}(x)$ is generalized Fourier density estimator in equation~\eqref{eq:generalized_Fourier_density_estimator} while $a_{n, R}(x)$ is defined as follows:
%\begin{align*}
%    a_{n, R}(x) : = \frac{R^{d}}{n A^{d}}\sum_{i = 1}^{n} Y_{i} \prod_{j = 1}^{d} \phi \left(\frac{\sin(R(x_{j} - X_{ij}))}{R(x_{j} - X_{ij})} \right).
%\end{align*}
The main difference between the generalized Fourier nonparametric regression estimator $f_{N,R}$ in equation~\eqref{eq:nonparametric_regression_Fourier} and the estimator $\widehat{f}_{\sigma}$ in equation~\eqref{eq:Gaussian_nonparametric_regression} is that the estimator $f_{N,R}$ utilizes the generalized Fourier density estimator to estimate the conditional distribution of the value vectors given the key vectors instead of the isotropic Gaussian kernel density estimator as in $\widehat{f}_{\sigma}$. As we highlighted in Section~\ref{sec:Fourierformer}, an important benefit of the generalized Fourier density estimator is that it can capture the complex dependencies of the features of the value vectors and the key vectors while the Gaussian kernel needs to have good covariance matrix to do that, which is computationally expensive in practice. 

We now have the following result establishing the mean square error (MSE) of $f_{N,R}$.
\begin{theorem}
\label{theorem:nonparametric_regression_first}
Assume that $\int_{\mathbb{R}} \phi \left(\frac{\sin(z)}{z}\right) z^{j} dz = 0$ for all $1 \leq j \leq m$ and $\int_{\mathbb{R}} \left|\phi \left(\frac{\sin(z)}{z}\right)\right| |z|^{j} dz < \infty$ for any $m + 1 \leq j \leq 2m + 2$ for some $m \in \mathbb{N}$. Then, for any $\bk \in \mathbb{R}^{D}$, there exist universal constants $C_{1}, C_{2}, C_{3}, C_{4}$ such that the following holds:
\begin{align*}
    \Exs \brackets{(f_{N,R}(\bk) - f(\bk))^2} \leq \parenth{\frac{C_{1}}{R^{2(m + 1)}} + \frac{(f(\bk) + C_{2}) R^{D}}{N}} \biggr/ \parenth{p^2(\bk) J(R)},
\end{align*}
where $J(R) = 1 - \frac{1}{p^2(\bk)} \parenth{\frac{C_{3}}{R^{2(m+1)}} + \frac{C_{4} R^{d} \log (N R)}{N}}$. Here, the outer expectation is taken with respect to the key vectors $\bk_{1}, \ldots, \bk_{N}$ and the noises $\varepsilon_{1}, \ldots, \varepsilon_{N}$.
\end{theorem}
Proof of Theorem~\ref{theorem:nonparametric_regression_first} is in Appendix~\ref{sec:proof:theorem:nonparametric_regression_first}. A few comments with Theorem~\ref{theorem:nonparametric_regression_first} are in order. First, by choosing $R$ to balance the bias and variance in the bound of the MSE of the nonparametric generalized Fourier estimator $f_{N,R}$, we have the optimal radius $R$ as $R = \mathcal{O}(N^{\frac{1}{2(m + 1) + D}})$. With that choice of the optimal radius $R$, the rate of $f_{N,R}$ is $\mathcal{O}(N^{-\frac{2(m+1)}{D + 2(m+1)}})$. Second, when $\phi(z) = z^{l}$ for $l \geq 6$, the assumption on the function $\phi$ of Theorem~\ref{theorem:nonparametric_regression_first} is satisfied with $m = 1$. Under this case, the rate of $f_{N,R}$ becomes $\mathcal{O}(N^{-\frac{4}{D + 4}})$. In Appendix~\ref{sec:additional_theory}, we also provide the rate of $f_{N,R}$ when $\phi(z) = z^{l}$ for some $l \leq 5$, which includes the original Fourier integral theorem. 
\subsubsection{FourierFormer}
\label{sec:fourierformer_definition}
Given the generalized Fourier nonparametric regression estimator $f_{N,R}$ in equation~\eqref{eq:nonparametric_regression_Fourier}, by plugging the query values $\bq_{1}, \ldots, \bq_{N}$ into that function, we obtain the following definition of the Fourier attention:
\begin{definition}[Fourier Attention]
\label{def:fish}
A Fourier attention is a multi-head attention that does nonparametric regression using the generalized Fourier nonparametric regression estimator $f_{N,R}$. The output $\hat{\bh}_i$ of the Fourier attention is then computed as
\begin{align}
    \hat{\bh}_i := f_{N,R}(\bq_{i}) = \frac{\sum_{i = 1}^{N} \bv_{i} \prod_{j = 1}^{D} \phi \left(\frac{\sin(R(q_{ij} - k_{ij}))}{R(q_{ij} - k_{ij})} \right)}{\sum_{i = 1}^{N} \prod_{j = 1}^{D} \phi \left(\frac{\sin(R(q_{ij} - k_{ij}))}{R(q_{ij} - k_{ij})} \right)} \quad \quad \forall \ i \in [N]. \label{eqn:fourier-attention}
\end{align}

\end{definition}

Given the Fourier Attention in Definition~\ref{def:fish}, we then give the definition of FourierFormer as follows.
\begin{definition}[FourierFormer]
A FourierFormer is a transformer that uses Fourier attention to capture dependency between tokens in the input sequence and the correlation between features in each token.
\end{definition}

\begin{remark}[The Nonnegativity of the Fourier Kernel]
\label{rm:nonneg-fourier}
The density estimation via generalized Fourier integral theorem in Section~\ref{sec:density_estimation_Fourier} does not require the generalized Fourier density estimator to be nonnegative. However, empirically, we observe that negative density estimator can cause instability in training the FourierFormer. Thus, in FourierFormer, we choose the function $\phi$ to be a nonnegative function to enforce the density estimator to be nonnegative. In particular, we choose $\phi$ to be power functions of the form $\phi(x) = x^{2m}$, where $m$ is an positive integer. Note that when $m=2$ and $m=4$, the kernels in our generalized Fourier integral estimators are the well-known Fejer-de la Vallee Poussin and Jackson-de la Vallee Poussin kernels~\cite{davis1975mean}.
\end{remark}

\subsection{An Efficient Implementation of the Fourier Attention} 
\label{sec:Implementation}
The Fourier kernel is implemented efficiently in the C++/CUDA extension developed by Pytorch \cite{NEURIPS2019_9015}. The idea is similar to the function \texttt{cdist} \cite{NEURIPS2019_9015}, which computes the p-norm distance between each pair of the two collections of row vectors. In our case, we aim to compute kernel functions that represent a Fourier attention in Definition~\ref{def:fish}. The core of this implementation is the following Fourier metric function $d_f$:
$$d_f(\bq_i , \bk_j) = \prod_{d=1}^{D} \phi\left(\frac{\sin(R(\bq_{id}-\bk_{jd})) } {  R(\bq_{id}-\bk_{jd}) }\right).$$
We directly implement $d_f$ as a \texttt{torch.autograd.Function} \cite{NEURIPS2019_9015} in which we provide an efficient way to compute forward and backward function ($d_f$ and gradient of $d_f$). While the implementation of the forward function is straight forward, the backward function is more tricky since we need to optimize the code to compute the gradient of $d_f$ w.r.t to variables $\bq$, $\bk$, and $R$ all at once. We can develop the backward function with highly parallel computation by exploiting GPU architecture and utilizing the reduction technique. The computational time is comparable to function \texttt{cdist}; thus, our FourierFormer implementation is as computationally time-efficient. 

\section{Experimental Results}\label{sec:experiments}
In this section, we numerically justify the advantage of FourierFormer over the baseline dot-product transformer on two large-scale tasks: language modeling on WikiText-103~\cite{DBLP:conf/iclr/MerityX0S17} (Section~\ref{subsec:wikitext}) and image classification on ImageNet~\cite{deng2009imagenet,russakovsky2015imagenet} (Section~\ref{subsec:ImageNet}). We aim to show that: (i) FourierFormer achieves better accuracy than the baseline transformer on a variety of practical tasks with different data modalities, and (ii) FourierFormer helps reduce head redundancy compared to the baseline transformer (Section~\ref{subsec:redundancy}). 

Throughout the section, we compare FourierFormers with the baseline dot-product transformers of the same configuration. In all experiments, we made the constant $R$ in Fourier attention (see equation~\eqref{eqn:fourier-attention}) to be a learnable scalar and set choose the function $\phi(x) = x^{4}$ (see Remark~\ref{rm:nonneg-fourier}). All of our results are averaged over 5 runs with different seeds. More details on the models and training are provided in Appendix~\ref{secapp:expdetails}. We also provide additional experimental results in Appendix~\ref{secapp:more-exp-results}.

\subsection{Language Modeling on WikiText-103}
\label{subsec:wikitext}
\textbf{Datasets and metrics} WikiText-103 is a collection of articles from Wikipedia, which have long contextual dependencies. The training set consists of about $28K$ articles containing $103M$ running words; this corresponds to text blocks of about 3600 words. The validation and test sets have $218K$ and $246K$ running words, respectively. Each of them contains $60$ articles and about $268K$ words. Our experiment follows the standard setting~\citep{DBLP:conf/iclr/MerityX0S17, schlag2021linear} and splits the training data into $L$-word independent long segments. For evaluation, we use a batch size of 1, and process the text sequence with a sliding window of size $L$. The last position is used for computing perplexity (PPL) except in the first segment, where all positions are evaluated as in~\citep{al2019character, schlag2021linear}.

\vspace{0.5 em}
\noindent
\textbf{Models and baselines:} Our implementation is based on the public code  by~\citep{schlag2021linear}.\footnote{Implementation available at \href{https://github.com/IDSIA/lmtool-fwp}{https://github.com/IDSIA/lmtool-fwp}.} We use their small and medium models in our experiments. In particular, for small models, the key, value, and query dimension are set to 128, and the training and evaluation context length are set to 256. For medium models, the key, value, and query dimension are set to 256, and the training and evaluation context length are set to 384. In both configurations, the number of heads is 8, the feed-forward layer dimension is 2048, and the number of layers is 16.

\vspace{0.5 em}
\noindent
\textbf{Results:} We report the validation and test perplexity (PPL) of FourierFormer versus the baseline transformer with the dot-product attention in Table~\ref{tab:lm-results}. FourierFormers attain much better PPL than the baselines in both small and medium configurations. For the small configuration, the improvements of FourierFormer over the baseline are 1.29 PPL in validation and 1.44 PPL in test. For the medium configuration, these improvements are 1.39 PPL in validation and 1.59 PPL in test. These results suggest that the advantage of FourierFormer over the baseline dot-product transformer grows with the model's size. This meets our expectation because  larger models has larger query and key dimensions, e.g. the language model with medium configuration in this experiment has the query and key dimension of 256 versus 128 as in the language model with small configuration. Since the advantage of FourierFormer results from the property that FourierFormer can capture correlation between features in query and key vectors, the larger the query and key dimensions are, the more advantage FourierFormer has.

\begin{table}[t!]
\small
    \caption{\small Perplexity (PPL) on WikiText-103 of FourierFormers compared to the baselines. FourierFormers achieve much better PPL than the baselines.}
    \vspace{0.1in}
    \begin{center}
    \scalebox{0.9}{\begin{tabular}{lcc}
    \toprule
        Method & Valid PPL & Test PPL \\
        \midrule
        {\it Baseline dot-product (small)} & 33.15  & 34.29 \\
        FourierFormer (small) & \bf 31.86  & \bf 32.85  \\
        \midrule\midrule
        {\it Baseline dot-product (medium)} & 27.90 & 29.60  \\
        FourierFormer (medium) & \bf 26.51  & \bf 28.01  \\
        \bottomrule
    \end{tabular}}
    \end{center}
\label{tab:lm-results}
\end{table}

\subsection{Image Classification on ImageNet}
\label{subsec:ImageNet}
\textbf{Datasets and metrics} The ImageNet dataset~\cite{deng2009imagenet,russakovsky2015imagenet} consists of $1.28M$ training images and $50K$ validation images. For this benchmark, the model learns to predict the category of the input image among 1000 categories. Top-1 and top-5 classification accuracies are reported.

\vspace{0.5 em}
\noindent
\textbf{Models and baselines:} We use the DeiT-tiny model~\cite{touvron2021training} with 12 transformer layers, 4 attention heads per layer, and the model dimension of 192. To train the models, we follow the same setting and configuration as for the baseline~\cite{touvron2021training}.\footnote{Implementation available at \href{https://github.com/facebookresearch/deit}{https://github.com/facebookresearch/deit}.}

\vspace{0.5 em}
\noindent
\textbf{Results:} 
We summarize our resuls in Table~\ref{tab:imagenet}. Same as in the language modeling experiment, for this image classification task, the Deit model equipped with FourierFormer significantly outperforms the baseline Deit dot-product transformer in both top-1 and top-5 accuracy. This result suggests that the advantage of FourierFormer over the baseline dot-product transformer holds across different data modalities.

\begin{table}[t!]
\small
    \caption{\small Top-1 and top-5 accuracy (\%) of FourierFormer Deit vs.\ the baseline Deit with dot-product attention. FourierFormer Deit outperforms the baseline in both top-1 and top-5 accuracy.}
    \vspace{0.1in}
    \begin{center}
    \scalebox{0.9}{\begin{tabular}{lcc}
    \toprule
        Method & Top-1 Acc & Top-5 Acc\\
        \midrule
        {\it Baseline DeiT} & 72.23 & 91.13\\
        FourierFormer DeiT & \bf 73.25 & \bf 91.66\\
        \bottomrule
    \end{tabular}}
    \end{center}

\label{tab:imagenet}
% \vspace{-0.1in}
\end{table}

\subsection{FourierFormer Helps Reducing Head Redundancy}
\label{subsec:redundancy}
To study the diversity between attention heads, given the model trained for the WikiText-103 language modeling task, we compute the average $\mathcal{L}_2$ distance between heads in each layer. We show the layer-average mean and variance of distances between heads in Table~\ref{tab:head_distance}. Results in Table~\ref{tab:head_distance} shows that FourierFormer obtains greater $\mathcal{L}_2$ distance between attention heads than the baseline transformer with the dot-product attention and thus helps reduce the head redundancy.
Note that we use the small configuration as specified in Section~\ref{subsec:wikitext} for both models. 
\begin{table}[t!]
\vspace{-0.1in}
\small
\caption{\small Laver-average mean and standard deviation of $\mathcal{L}_2$ distances between heads of FourierFormer versus the baseline transformer with dot-product attention trained for the WikiText-103 language modeling task. FourierFormer has greater $\mathcal{L}_2$ distance between heads than the baseline and thus captures more diverse attention patterns.}
\vspace{0.1in}
    \centering
    \scalebox{0.9}{\begin{tabular}{lcc}
    \toprule
    Method & Train & Test 
    \\
    \midrule
        {\it Baseline dot-product} & $6.20 \pm 2.30$ & $6.17 \pm 2.30$  \\
        FourierFormer & $\bf 7.45 \pm 2.50$ & $\bf 7.37 \pm 2.44$ \\
         \bottomrule
    \end{tabular}}
    \label{tab:head_distance}
\end{table}

\section{Related Work}\label{sec:related_work}
{\bf Interpretation of Attention Mechanism in Transformers:}
Recent works have tried to gain an understanding of transformer's attention from different perspectives.~\cite{tsai2019transformer} considers attention as applying kernel smoother over the inputs. Extending this kernel approach,~\cite{katharopoulos2020transformers,choromanski2021rethinking,wang2020linformer} linearize the softmax kernel in dot-product attention and propose a family of efficient transformers with linear computational and memory complexity.~\cite{cao2021choose} then shows that these linear transformers are comparable to a Petrov-Galerkin
projection~\cite{reddy2004introduction}, suggesting that the softmax normalization in the dot-product attention is sufficient but not necessary. Other works provide an understanding of attention in transformers via ordinary/partial differential equation include~\cite{lu2019understanding,sander2022sinkformers}.  In addition,~\cite{tang2021probabilistic,gabbur2021probabilistic,zhang-feng-2021-modeling-concentrated, Tam_mixturekeys} relate attentions in transformers to a Gaussian mixture models. Several works also connect the attention mechanism to graph-structured learning and message passing in graphical models~\cite{wang2018non,shaw-etal-2018-self,kreuzer2021rethinking}. Our work focuses on deriving the connection between self-attention and nonparametric kernel regression and exploring better regression estimator, such as the generalized Fourier nonparametric regression estimator, to improve the performance of transformers.

\vspace{0.5 em}
\noindent
{\bf Redundancy in Transformers:}
\cite{dalvi2020analyzing,  NEURIPS2019_2c601ad9, durrani2020analyzing} show that neurons and attention heads in the pre-trained transformer are redundant and can be removed when applied on a downstream task. By studying the contextualized embeddings in pre-trained networks, it has been demonstrated that the learned representations from these redundant models are highly anisotropic~\citep{mu2018allbutthetop,ethayarajh2019contextual}. Furthermore,~\cite{sanh2019distilbert, sun2019patient, voita2019analyzing, sajjad2020poor} employ knowledge distillation and sparse approximation to enhance the efficiency of transformers. Our FourierFormer is complementary to these methods and can be combined with them.

\section{Concluding Remarks}\label{sec:conclusion}
In this paper, we establish the correspondence between the nonparametric kernel regression and the self-attention in transformer. We then develop the generalized Fourier integral estimators and propose the FourierFormer, a novel class of transformers that use the generalized Fourier integral estimators to construct their attentions for efficiently capturing the correlations between features in the query and key vectors. We theoretically prove the approximation guarantees of the generalized Fourier integral estimators and empirically validate the advantage of FourierFormer over the baseline transformer with the dot-product attention in terms of accuracy and head redundancy reduction. It is interesting to incorporate robust kernels into the nonparametric regression framework of FourierFormer to enhance the robustness of the model under data perturbation and adversarial attacks. A limitation of FourierFormer is that it still has the same quadratic computational and memory complexity as the baseline transformer with the dot-product attention. We leave the development of the linear version of FourierFormer that achieves linear computational and memory complexity as future work. It is worth noting that there is no potential negative societal impacts of FourierFormer.

% as the baseline transformer, our previous network that uses cdist.

\appendix
\begin{center}
{\bf \Large{Supplement to ``FourierFormer: Transformer Meets Generalized Fourier Integral Theorem"}}
\end{center}
In the supplementary material, we collect proofs, additional theories, and experiment results deferred from the main text. In Appendix~\ref{sec:additional_theory}, we provide additional theoretical results for generalized Fourier density estimator and for generalized Fourier nonparametric regression estimator. We provide proofs of key results in the main text and additional theories in Appendix~\ref{sec:Proofs}. We present experiment details in Appendix~\ref{secapp:expdetails} while including additional experimental results in Appendix~\ref{secapp:more-exp-results}.
%\section{Comparing to Gaussian Kernel}
%\label{sec:compare_Gaussian_kernel}
\section{Additional Theoretical Results}
\label{sec:additional_theory}
In this section, we provide additional theoretical results for generalized Fourier density estimator in Appendix~\ref{sec:additional_theory_Generalized Fourier density estimator} and for generalized Fourier nonparametric regression estimator in Appendix~\ref{sec:additional_theory_generalized_Fourier_nonparametric_regression}.
\subsection{Generalized Fourier density estimator}
\label{sec:additional_theory_Generalized Fourier density estimator}
We now establish the MISE rate of $p_{N,R}^{\phi}$ in equation~\eqref{eq:generalized_Fourier_density_estimator} when $\phi(z) = z^{l}$ and $l \in \{1,2\}$. We consider the following tail bounds on the Fourier transform of the true density function $p$ as follows.
\begin{definition}
\label{definition:smoothness}
(1) We say that $p$ is \emph{supersmooth} of order $\alpha$ if we have universal constants $C_{1}$ and $C_{2}$ such that the following inequalities hold for almost surely $x \in \mathbb{R}^{D}$:
\begin{align*}
\abss{ \widehat{p}(x)} & \leq C_{1} \exp \parenth{ -C_{2} \parenth{ \sum_{j = 1}^{D} |x_{j}|^{\alpha}} }.
\end{align*}
Here, $\widehat{p}$ denotes the Fourier transform of the function $p$.

\noindent
(2) The function $p$ is \emph{ordinary smooth} of order $\beta$ if there exists universal constant $c$ such that the following inequality holds for almost surely $x \in \mathbb{R}^{D}$:
\begin{align*}
    \abss{ \widehat{p}(x)} & \leq  c \cdot \prod_{j = 1}^{D} \frac{1}{(1 + |x_{j}|^{\beta})}.
\end{align*}
\end{definition}
The notions of supersmoothness and ordinary smoothness had been used widely in deconvolution problems~\cite{Fan-91} and density estimation problems~\cite{davis1975mean, Tsy09, Ho21}. The supersmooth condition is satisfied when the function $p$ is Gaussian distribution or Cauchy distribution while the ordinary smooth condition is satisfied when the function $p$ is Laplace distribution and Beta distribution. 

Based on the smoothness conditions in Definition~\ref{definition:smoothness}, we have the following result regarding the mean-square integrated error (MISE) of the function generalized Fourier density estimator~\eqref{eq:generalized_Fourier_density_estimator} (see equation~\eqref{eq:MISE} for a definition of MISE) when $\phi(z) = z^{l}$ and $l \in \{1, 2\}$. 
\begin{theorem}
\label{theorem:density_estimation_second}
(a) When $\phi(z) = z$, the following holds:
\begin{itemize}
    \item (Supersmooth setting) If the true density function $p$ is supersmooth function of order $\alpha$ for some $\alpha > 0$, then there exists universal constants $\bar{C}_{1}, \bar{C}_{2},$ and $\bar{C}_{3}$ such that as long as $R \geq \bar{C}_{1}$ we have
    \begin{align*}
        \text{MISE}(p_{N,R}^{\phi}) \leq \bar{C}_{2} \left(\radius^{\max \{1 - \alpha, 0\}} \exp(-\bar{C}_{3} \radius^{\alpha}) + \frac{R^{D}}{N} \right).
    \end{align*}
    \item (Ordinary smooth setting) If the true density function $p$ is ordinary smooth function of order $\beta$ for some $\beta > 1$, then there exists universal constants $\bar{c}$ such that
    \begin{align*}
    \text{MISE}(p_{N,R}^{\phi}) \leq \bar{c} \left(\radius^{-\beta + 1} + \frac{R^{D}}{N} \right).
    \end{align*}
\end{itemize}
(b) When $\phi(z) = z^2$, the following holds
\begin{itemize}
    \item (Supersmooth setting) If the true density function $p$ is supersmooth function of order $\alpha$ for some $\alpha > 0$, then there exists universal constants $C_{1}'$ and $C_{2}'$ such that as long as $R \geq C_{1}'$ we have
    \begin{align*}
        \text{MISE}(p_{N,R}^{\phi}) \leq C_{2}' \parenth{\frac{1}{R^2} + \frac{R^{D}}{N}}.
    \end{align*}
    \item (Ordinary smooth setting) If the true density function $p$ is ordinary smooth function of order $\beta$ for some $\beta > 3$, then there exists universal constants $c'$ such that
    \begin{align*}
        \text{MISE}(p_{N,R}^{\phi}) \leq c' \parenth{\frac{1}{R^2} + \frac{R^{D}}{N}}.
    \end{align*}
\end{itemize}
\end{theorem}
Proof of Theorem~\ref{theorem:density_estimation_second} is in Appendix~\ref{subsec:proof:theorem:density_estimation_second}. A few comments with the results of Theorem~\ref{theorem:density_estimation_second} are in order. 

\vspace{0.5 em}
\noindent
\textbf{When $\phi(z) = z$:} As part (a) of Theorem~\ref{theorem:density_estimation_second} indicates, when the function $p$ is supersmooth, by choosing the radius $R$ to balance the bias and variance, we have the optimal $R$ as $R = \left(\frac{\log(N)}{\bar{C}_{3}}\right)^{1/\alpha}$ and the MISE rate of the generalized Fourier density estimator $p_{N,R}^{\phi}$ becomes $\mathcal{O} \left( \frac{\log(N)^{D/\alpha}}{N} \right)$. It indicates that, the MISE rate of $p_{N,R}^{\phi}$ is parametric when the function $p$ is supersmooth. On the other hand, when the function $p$ is ordinary smooth, the optimal $R$ becomes $\mathcal{R} = \mathcal{O}(N^{\frac{1}{D + \beta - 1}})$ and the MISE rate becomes $\mathcal{O} \left(N^{- \frac{\beta - 1}{D + \beta - 1}} \right)$. It is slower than the MISE rate when the function $p$ is supersmooth.

\vspace{0.5 em}
\noindent
\textbf{When $\phi(z) = z^2$:} The results of part (b) of Theorem~\ref{theorem:density_estimation_second} demonstrate that the upper bounds for the MISE rate of the generalized Fourier density estimator $p_{N,R}^{\phi}$ is similar for both the supersmooth and ordinary smooth settings. The optimal radius $R = \mathcal{O}\parenth{N^{\frac{1}{D + 2}}}$ and the MISE rate of the estimator is $\mathcal{O}\parenth{N^{-\frac{2}{D + 2}}}$.
\subsection{Generalized Fourier nonparametric regression estimator}
\label{sec:additional_theory_generalized_Fourier_nonparametric_regression}
In this appendix, we provide additional result for the mean square error (MSE) rate of the generalized Fourier nonparametric regression estimator $f_{N,R}$ in equation~\eqref{eq:nonparametric_regression_Fourier} when $\phi(z) = z$, namely, the setting of the Fourier integral theorem. The results when $\phi(z) = z^{l}$ for $l \in \{2,3,4,5\}$ are left for the future work. 

When $\phi(z) = z$, the MSE rate of $f_{N,R}$ had been established in Theorem 9 of Ho et al.~\cite{Ho21} when the function $p$ is supersmooth function. Here, we restate that result for the completeness.
\begin{theorem}
\label{theorem:nonparametric_Fourier_regression_second}
Assume that the function $p$ is supersmooth function of order $\alpha$ for some $\alpha > 0$ and $\sup_{\bk \in \mathbb{R}^{D}} |p(\bk)| < \infty$. Furthermore, we assume that the function $f$ in the nonparametric regression model~\eqref{eq:nonparametric_regression_softmax_former} is such that $\sup_{\bk \in \mathbb{R}^{D}} |f^{2}(\bk) p(\bk)| < \infty$ and 
\begin{align*}
    |\widehat{f.p}(\bt)| \leq C_{1} Q(|t_{1}|, |t_{2}|, \ldots, |t_{D}|) \exp \parenth{-C_{2} \parenth{\sum_{j = 1}^{D} |t_{j}|^{\alpha}}},
\end{align*}
where $\widehat{f.p}(\bt)$ is the Fourier transform of the function $f.p$, $C_{1}$ and $C_{2}$ are some universal constants, and $Q(|t_{1}|, |t_{2}|, \ldots, |t_{D}|)$ is some polynomial function of $|t_{1}|, \ldots, |t_{D}|$ with non-negative coefficients. Then, we can find universal constants $C_{3}, C_{4}, C_{5}$ such that as long as $R \geq C_{3}$ we have
\begin{align*}
    \Exs \brackets{(f_{N,R}(\bk) - f(\bk))^2} \leq C_{4} \frac{R^{\max\{2 \text{deg}(Q) + 2 - 2\alpha, 0 \}} \exp \parenth{- 2 C_{2} R^{\alpha}} + \frac{(f(\bk) + C_{5}) R^{D}}{N}}{p^2(\bk) \bar{J}(R)},
\end{align*}
where $\text{deg}(Q)$ denotes the degree of the polynomial function $Q$, and we define $\bar{J}(R) = 1 - \frac{R^{\max \{2 - 2 \alpha, 0\}} \exp ( - 2 C_{2} R^{\alpha}) + \frac{R^{D} \log(NR)}{N}}{p^{2}(\bk)}$. 
\end{theorem}
Proof of Theorem~\ref{theorem:nonparametric_Fourier_regression_second} is similar to the proof of Theorem 9 of Ho et al.~\cite{Ho21}; therefore, it is omitted. The result of Theorem~\ref{theorem:nonparametric_Fourier_regression_second} indicates that the optimal radius $R = \parenth{\frac{\log(N)}{2 C_{2}}}^{1/\alpha}$ and the MSE rate of the generalized Fourier nonparametric regression estimator $f_{N,R}$ is $\mathcal{O} \parenth{\frac{\log(N)^{D/\alpha}}{N}}$. 
\section{Proofs} \label{sec:Proofs}
In this Appendix, we provide proofs for key results in the paper and in Appendix~\ref{sec:additional_theory}.
\subsection{Proof of Theorem~\ref{theorem:density_estimation_first}}
\label{subsec:proof:theorem:density_estimation_first}
Recall that, $\bk_{1}, \bk_{2}, \ldots, \bk_{N} \in \mathbb{R}^{D}$ are i.i.d. samples from the density function $p$. In equation~\eqref{eq:generalized_Fourier_density_estimator}, the generalized Fourier density estimator of $p_{0}$ is given by:
\begin{align*}
    p_{N, R}^{\phi}(\bk) = \frac{R^{D}}{N A^{D}} \sum_{i = 1}^{N} \prod_{j = 1}^{D} \phi \left(\frac{\sin(R(k_{j} - k_{ij}))}{R(k_{j} - k_{ij})} \right),
\end{align*}
where $A = \int_{\mathbb{R}} \phi \left(\frac{\sin(z)}{z}\right) dz$, $\bk_{i} = (k_{i1}, \ldots, k_{iD})$, and $\bk = (k_{1}, \ldots, k_{D})$. Direct calculation demonstrates that
\begin{align}
    \mathbb{E} [p_{N,R}^{\phi}(\bk)] & = \frac{R^{D}}{A^{D}} \int_{\mathbb{R}^{D}}  \prod_{j = 1}^{D} \phi \left(\frac{\sin(R(k_{j} - y_{j}))}{R(k_{j} - y_{j})} \right) p(\by) d\by \nonumber \\
    & = \frac{1}{A^{D}} \int_{\mathbb{R}^{D}} \prod_{j = 1}^{D} \phi \left(\frac{\sin(y_{j})}{y_{j}} \right) p \left(\bk - \frac{\by}{R} \right) d\by. \label{eq:key_equation_expectation}
\end{align}
An application of Taylor expansion up to the $m$-th order indicates that
\begin{align}
    p \left(\bk - \frac{\by}{R} \right) = \sum_{0 \leq |\alpha| \leq m} \frac{1}{R^{|\alpha|} \alpha!} \prod_{j = 1}^{D} (-y_{j})^{\alpha_{j}} \dfrac{\partial^{|\alpha|} p}{\partial{\bk^{\alpha}}} (\bk) + \bar{R}(\bk, \by), \label{eq:key_equation_expectation_first}
\end{align}
where $\alpha = (\alpha_{1}, \ldots, \alpha_{d})$, $|\alpha| = \sum_{j = 1}^{d} \alpha_{j}$, and $\bar{R}(\bk, \by)$ is Taylor remainder admitting the following form:
\begin{align}
    \bar{R}(\bk, \by) = \sum_{|\beta| = m + 1} \frac{m + 1}{R^{m + 1}\beta!} \prod_{j = 1}^{D} (-y_{j})^{\beta_{j}} \int_{0}^{1} (1 - t)^{m} \dfrac{\partial^{m + 1} p}{\partial{\bk^{\beta}}} \left(\bk - \frac{t \by}{R} \right) dt. \label{eq:key_equation_expectation_second}
\end{align}
Plugging equations~\eqref{eq:key_equation_expectation_first} and~\eqref{eq:key_equation_expectation_second} into equation~\eqref{eq:key_equation_expectation}, we find that
\begin{align*}
    & \mathbb{E} [p_{N,R}^{\phi}(\bk)] \\
    & = p(\bk) + \frac{1}{A^{D}} \sum_{1 \leq |\alpha| \leq m} \frac{1}{R^{|\alpha|} \alpha!} \int_{\mathbb{R}^{D}} \prod_{j = 1}^{D} \phi \left(\frac{\sin(y_{j})}{y_{j}} \right) \prod_{j = 1}^{d} (-y_{j})^{\alpha_{j}} \dfrac{\partial^{|\alpha|} p}{\partial{\bk^{\alpha}}} (\bk) d\by \\
    & + \frac{1}{A^{D}} \sum_{|\beta| = m + 1} \frac{m + 1}{R^{m + 1}\beta!} \int_{\mathbb{R}^{D}} \prod_{j = 1}^{D} \phi \left(\frac{\sin(y_{j})}{y_{j}} \right) \prod_{j = 1}^{D} (-y_{j})^{\beta_{j}} \int_{0}^{1} (1 - t)^{m} \dfrac{\partial^{m + 1} p_{0}}{\partial{\bk^{\beta}}} \left(\bk - \frac{t \by}{R} \right) d\by dt. 
\end{align*}
According to the hypothesis that $\int_{\mathbb{R}} \phi \left(\frac{\sin(z)}{z}\right) z^{j} dz = 0$ for all $1 \leq j \leq m$, we obtain that
\begin{align*}
    \int_{\mathbb{R}^{D}} \prod_{j = 1}^{D} \phi \left(\frac{\sin(y_{j})}{y_{j}} \right) \prod_{j = 1}^{D} (-y_{j})^{\alpha_{j}} \dfrac{\partial^{|\alpha|} p}{\partial{\bk^{\alpha}}} (\bk) d\by = 0
\end{align*}
for any $\alpha = (\alpha_{1}, \ldots, \alpha_{d})$ such that $1 \leq |\alpha| \leq m$. Collecting the above results, we arrive at
\begin{align*}
    & |\mathbb{E} [p_{N,R}^{\phi}(\bk)] - p(\bk)| \\
    & = \left| \frac{1}{A^{D}} \sum_{|\beta| = m + 1} \frac{m + 1}{R^{m + 1}\beta!} \int_{\mathbb{R}^{D}} \prod_{j = 1}^{D} \phi \left(\frac{\sin(y_{j})}{y_{j}} \right) \prod_{j = 1}^{D} (-y_{j})^{\beta_{j}} \int_{0}^{1} (1 - t)^{m} \dfrac{\partial^{m + 1} p}{\partial{\bk^{\beta}}} \left(\bk - \frac{t \by}{R} \right) d\by dt \right| \\
    & \leq \frac{1}{A^{D}} \sum_{|\beta| = m + 1} \frac{m + 1}{R^{m + 1}\beta!} \int_{\mathbb{R}^{D}} \prod_{j = 1}^{D} \left| \phi \left(\frac{\sin(y_{j})}{y_{j}} \right) \right| \prod_{j = 1}^{D} |y_{j}|^{\beta_{j}} \int_{0}^{1} (1 - t)^{m} \left|\dfrac{\partial^{m + 1} p}{\partial{\bk^{\beta}}} \left(\bk - \frac{t \by}{R} \right)\right| d\by dt.
\end{align*}
Since the function $p \in \mathcal{C}^{m+1}(\mathbb{R}^{D})$, we can find positive constant $M$ such that $\|\dfrac{\partial^{m + 1} p}{\partial{\bk^{\beta}}}(\bk) \|_{\infty} \leq M$ for all $\beta = (\beta_{1}, \ldots, \beta_{d})$ such that $|\beta| = m + 1$. Therefore, we find that
\begin{align*}
    |\mathbb{E} [p_{N,R}^{\phi}(\bk)] - p(\bk)| & \leq \frac{M}{A^{D}} \sum_{|\beta| = m + 1} \frac{m + 1}{R^{m + 1}\beta!} \int_{\mathbb{R}^{D}} \prod_{j = 1}^{D} \left| \phi \left(\frac{\sin(y_{j})}{y_{j}} \right) \right| \prod_{j = 1}^{D} |y_{j}|^{\beta_{j}} d\by \int_{0}^{1} (1 - t)^{m} dt \\
    & = \frac{M}{A^{D}} \sum_{|\beta| = m + 1} \frac{1}{R^{m + 1}\beta!} \int_{\mathbb{R}^{D}} \prod_{j = 1}^{D} \left| \phi \left(\frac{\sin(y_{j})}{y_{j}} \right) \right| \prod_{j = 1}^{D} |y_{j}|^{\beta_{j}} d\by.
\end{align*}
For any $\beta = (\beta_{1}, \ldots, \beta_{D})$ such that $|\beta| = m + 1$, an application of the AM-GM inequality indicates that $\prod_{j = 1}^{D} |y_{j}|^{\beta_{j}} \leq m (\sum_{j = 1}^{D} |y_{j}|^{m + 1})$. Hence, putting these results together leads to
\begin{align*}
    |\mathbb{E} [p_{N,R}^{\phi}(\bk)] - p(\bk)| \leq \frac{M m}{A^{D} R^{m + 1}} \sum_{|\beta| = m + 1} \frac{1}{\beta!} \int_{\mathbb{R}^{D}} \prod_{j = 1}^{D} \left| \phi \left(\frac{\sin(y_{j})}{y_{j}} \right) \right| \left(\sum_{j = 1}^{D} |y_{j}|^{m + 1} \right) d\by.
\end{align*}
From the hypothesis, we have $\int_{\mathbb{R}} \left|\phi \left(\frac{\sin(z)}{z}\right)\right| |z|^{m + 1} dz < \infty$. As a consequence, we can find a universal constant $C$ depending on $A$ and $d$ such that
\begin{align*}
    |\mathbb{E} [p_{n,R}^{\phi}(\bk)] - p(\bk)| \leq \frac{C}{R^{m+1}}
\end{align*}
for all $\bk \in \mathbb{R}^{D}$. 

\textbf{Bounding the variance:} We now move to bound the variance of $p_{N, R}^{\phi}(\bk)$. Indeed, direct computation indicates that
\begin{align*}
    \text{Var}[p_{N, R}^{\phi}(\bk)] & = \frac{R^{2D}}{n A^{2D}} \text{Var} \left[\prod_{j = 1}^{D} \phi \left(\frac{\sin(R(k_{j} - K_{.j}))}{R(x_{j} - K_{.j})} \right)\right] \\
    & \leq \frac{R^{2D}}{n A^{2D}} \mathbb{E} \left[\prod_{j = 1}^{D} \phi^2 \left(\frac{\sin(R(k_{j} - K_{.j}))}{R(k_{j} - K_{.j})} \right)\right] \\
    & = \frac{R^{D}}{n A^{2D}} \int_{\mathbb{R}^{D}} \prod_{j = 1}^{D} \phi^2 \left(\frac{\sin(y_{j}))}{y_{j}}\right) p \left(\bk - \frac{\by}{R} \right) d\by \leq \frac{R^{D} \|p \|_{\infty}}{N A^{2D}} \int_{\mathbb{R}^{D}} \prod_{j = 1}^{D} \phi^2 \left(\frac{\sin(y_{j}))}{y_{j}}\right) d\by
\end{align*}
where the variance and the expectation are taken with respect to $K = (K_{.1}, \ldots, K_{.d}) \sim p$. As $\int_{\mathbb{R}} \phi^2 \left(\frac{\sin(z))}{z}\right) dz < \infty$, there exists a universal constant $C'$ depending on $A$ and $D$ such that
\begin{align*}
    \text{Var}[p_{N, R}^{\phi}(\bk)] \leq \frac{C' R^{D}}{N}.
\end{align*}
As a consequence, we obtain the conclusion of the theorem. 
\subsection{Proof of Theorem~\ref{theorem:density_estimation_second}}
\label{subsec:proof:theorem:density_estimation_second}
From the Plancherel theorem, we obtain that
\begin{align}
    \int_{\mathbb{R}^{D}} \left[(p_{N,R}^{\phi}(\bk) - p(\bk) \right]^2 d\bk = \frac{1}{(2\pi)^{D}} \int_{\mathbb{R}^{D}} \left[\widehat{p}_{N,R}^{\phi}(\bt) - \widehat{p}(\bt)\right]^2 d\bt, \label{eq:Plancherel_theorem}
\end{align}
where $\widehat{p}_{N,R}^{\phi}$ and $\widehat{p}$ are respectively the Fourier transforms of $p_{N,R}$ and $p$. From the definition of generalized Fourier density estimator $p_{N,R}^{\phi}$ in equation~\eqref{eq:generalized_Fourier_density_estimator}, it is clear that
\begin{align*}
    \widehat{p}_{N,R}^{\phi}(t) = \frac{1}{N} \sum_{i = 1}^{N} \exp(i \bt^{\top} \bk_{i}) \prod_{j = 1}^{D} K_{R}(t_{j}),
\end{align*}
for any $\bt = (t_{1}, \ldots, t_{D}) \in \mathbb{R}^{D}$ where we define $K_{R}(y) : = \frac{1}{\pi} \int_{\mathbb{R}} R \phi \left(\frac{\sin(R x)}{R x} \right) \exp(i y x) dx$ for any $y \in \mathbb{R}$. To ease the presentation, we denote $\bar{K}_{R}(\bt) : = \prod_{j = 1}^{D} K_{R}(t_{j})$ and $\varphi_{N}(\bt) = \frac{1}{N} \sum_{i = 1}^{N} \exp(i \bt^{\top} \bk_{i})$ for any $\bt = (t_{1}, t_{2}, \ldots, t_{D}) \in \mathbb{R}^{D}$. Based on these notations, we can rewrite 
$$\widehat{p}_{N,R}^{\phi}(\bt) = \varphi_{N}(\bt) \bar{K}_{R}(\bt)$$ 
Direct calculation shows that $\mathbb{E}_{\bk_{1}^{N}} [\varphi_{N}(\bt)] = \widehat{p}(\bt)$ for any $\bt \in \mathbb{R}^{D}$ where $\bk_{1}^{N} : = (\bk_{1}, \ldots, \bk_{n})$. Furthermore, we have
\begin{align*}
    \mathbb{E}_{\bk_{1}^{N}} [|\varphi_{N}(\bt)|^2] = \mathbb{E}[\varphi_{N}(\bt)\varphi_{N}(-\bt)] & = \mathbb{E} \left[\left(\frac{1}{N} \sum_{i = 1}^{N} \exp(i \bt^{\top} \bk_{i})\right) \left(\frac{1}{N} \sum_{i = 1}^{N} \exp(- i \bt^{\top} \bk_{i})\right)\right] \\
    & = \frac{1}{N} + \frac{(N - 1)}{N} \mathbb{E} \left[\exp(i \bt^{\top} \bk) \exp(- i \bt^{\top} \bk)\right] \\
    & = \frac{1}{N} + \frac{(N - 1)}{N} |\widehat{p}(\bt)|^2.
\end{align*}
Collecting the above results, we have the following equations:
\begin{align}
    \mathbb{E}_{\bk_{1}^{n}} \left[\int_{\mathbb{R}^{D}} \left[\widehat{p}_{N,R}^{\phi}(\bt) - \widehat{p}(\bt)\right]^2 d\bt \right] & = \mathbb{E}_{\bk_{1}^{n}} \left[\int_{\mathbb{R}^{D}} \left[\varphi_{N}(\bt) \bar{K}_{R}(\bt) - \widehat{p}(\bt)\right]^2 d\bt \right] \nonumber \\
    & = \mathbb{E}_{\bk_{1}^{n}} \left[ \int_{\mathbb{R}^{D}} \left[(\varphi_{n}(\bt) - \widehat{p}(\bt))\bar{K}_{R}(\bt) - \widehat{p}(\bt)(1 - \bar{K}_{R}(\bt))\right]^2 d\bt \right] \nonumber \\
    & = \int_{\mathbb{R}^{D}} \mathbb{E}_{\bk_{1}^{N}} \left[(\varphi_{N}(\bt) - \widehat{p}(\bt))^2\right] \bar{K}_{R}^2(\bt) + \widehat{p}^2(\bt)(1 - \bar{K}_{R}(\bt))^2 d\bt \nonumber \\
    & = \int_{\mathbb{R}^{D}} \widehat{p}^2(\bt)(1 - \bar{K}_{R}(\bt))^2 d\bt + \frac{1}{N} \int_{\mathbb{R}^{D}} (1 - |\widehat{p}(\bt)|^2) \bar{K}_{R}^2(\bt) d\bt. \label{eq:bias_variance_tradeoff}
\end{align}
Combining the results from equations~\eqref{eq:Plancherel_theorem} and~\eqref{eq:bias_variance_tradeoff}, we find that
\begin{align}
    \text{MISE}(p_{N,R}^{\phi}) & = \mathbb{E}_{\bk_{1}^{N}} \left[\int_{\mathbb{R}^{D}} \left[(p_{N,R}^{\phi}(\bk) - p(\bk) \right]^2 d\bk \right] \nonumber \\
    & = \frac{1}{(2\pi)^{D}} \left( \int_{\mathbb{R}^{D}} \widehat{p}^2(\bt)(1 - \bar{K}_{R}(\bt))^2 d\bt + \frac{1}{N} \int_{\mathbb{R}^{D}} (1 - |\widehat{p}(\bt)|^2) \bar{K}_{R}^2(\bt) d\bt \right). \label{eq:key_equation}
\end{align}
\subsubsection{When $\phi(z) = z$}
We first consider the setting when $\phi(z) = z$, namely, the setting of the Fourier integral theorem. Under this setting, direct computation indicates that 
\begin{align*}
    \bar{K}_{R}(\bt) = \prod_{i = 1}^{d} \bold{1}_{\{|t_{i}| \leq R\}}. 
\end{align*}
Given the smoothness assumptions on the function $p$, we have two settings on that function.

\textbf{Supersmooth setting of the function $p$:} When the function $p$ is supersmooth density, we have
\begin{align*}
\abss{ \widehat{p}(\bt)} & \leq C_{1} \exp \parenth{ -C_{2} \parenth{ \sum_{j = 1}^{D} |t_{j}|^{\alpha}} },
\end{align*}
where $C_{1}$ and $C_{2}$ are some universal constants. Therefore, we find that
\begin{align}
    \int_{\mathbb{R}^{D}} \widehat{p}^2(\bt)(1 - \bar{K}_{R}(\bt))^2 d\bt = \int_{\mathbb{R}^{D} \backslash [-R,R]^{D}} \widehat{p}^2(\bt) d\bt & \leq C_{1} \int_{\mathbb{R}^{D} \backslash [-R,R]^{D}} \exp \parenth{ -C_{2} \parenth{ \sum_{j = 1}^{D} |t_{j}|^{\alpha}} } d\bt \nonumber \\
    & \leq C_{1} \sum_{i = 1}^{D} \int_{B_{i}} \exp \parenth{ -C_{2} \parenth{ \sum_{j = 1}^{D} |t_{j}|^{\alpha}} } d\bt, \label{eq:key_supersmooth_first}
\end{align}
where $B_{i} : = \{\bt \in \mathbb{R}^{D}: \ |t_{i}| \geq R\}$. We now proceed to bound $\int_{B_{i}} \exp \parenth{ -C_{2} \parenth{ \sum_{j = 1}^{D} |t_{j}|^{\alpha}} } d\bt$ for all $i \in [D]$. Indeed, we have that
\begin{align*}
    \int_{B_{i}} \exp \parenth{ -C_{2} \parenth{ \sum_{j = 1}^{D} |t_{j}|^{\alpha}} } d\bt & = \parenth{ \int_{\mathbb{R}} \exp(-C_{2} |x|^{\alpha})dx}^{D - 1} \cdot \int_{|x| \geq \radius} \exp(-C_{2} |x|^{\alpha})dx \\
    & = \frac{C_{2} \alpha^{D - 1}}{\parenth{2 C_{2}\Gamma(1/ \alpha)}^{D - 1}} \cdot \int_{|x| \geq \radius} \exp(-C_{2} |x|^{\alpha})dx.
\end{align*}
When $\alpha \geq 1$, we have that 
\begin{align*}
\int_{R}^{\infty} \exp \parenth{-C_{2} x^{\alpha}} dx \leq \int_{\radius}^{\infty} x^{\alpha - 1} \exp \parenth{-C_{2} x^{\alpha}} dx = \exp(-C_{2} \radius^{\alpha})/ (C_{2} \alpha).
\end{align*}
When $\alpha \in (0, 1)$, then we find that
\begin{align*}
    \int_{\radius}^{\infty} \exp(-C_{2} x^{\alpha}) dx & = \int_{\radius}^{\infty} x^{1 - \alpha} x^{\alpha - 1} \exp(-C_{2} x^{\alpha}) dx \\
    & \leq \frac{\radius^{1 - \alpha} \exp \parenth{-C_{2}\radius^{\alpha}}}{C_{2} \alpha} + \frac{1 - \alpha}{C_{2} \alpha \radius^{\alpha}} \int_{\radius}^{\infty} \exp(-C_{2} x^{\alpha}) dx,
\end{align*}
When the $\radius$ is such that $\radius^{\alpha} \geq \frac{2(1 - \alpha)}{C_{2} \alpha}$, the above inequality becomes
\begin{align*}
    \int_{\radius}^{\infty} \exp(-C_{2} x^{\alpha}) dx \leq \frac{2 \radius^{1 - \alpha} \exp \parenth{-C_{2}\radius^{\alpha}}}{C_{2} \alpha}.
\end{align*}
Collecting the above results, we arrive at
\begin{align}
    \int_{|x| \geq \radius} \exp(-C_{2} |x|^{\alpha})dx \leq \frac{4 \radius^{\max \{1 - \alpha, 0\}}}{C_{2} \alpha} \exp(-C_{2}\radius^{\alpha}). \label{eq:key_supersmooth_second}
\end{align}
Plugging the inequality~\eqref{eq:key_supersmooth_second} into the inequality~\eqref{eq:key_supersmooth_first}, there exists universal constant $C_{3}$ depending on $\alpha$ and $D$ such that
\begin{align}
    \int_{\mathbb{R}^{D}} \widehat{p}^2(\bt)(1 - \bar{K}_{R}(\bt))^2 d\bt \leq C_{3} \radius^{\max \{1 - \alpha, 0\}} \exp(-C_{1} \radius^{\alpha}). \label{eq:upper_bound_supersmooth_bias}
\end{align}
On the other hand, we also have
\begin{align}
    \frac{1}{N} \int_{\mathbb{R}^{D}} (1 - |\widehat{p}(\bt)|^2) \bar{K}_{R}^2(\bt) d\bt \leq \frac{1}{N} \int_{\mathbb{R}^{D}} \bar{K}_{R}^2(\bt) \leq \frac{R^{D}}{N}. \label{eq:upper_bound_supersmooth_variance}
\end{align}
Combining the results from equations~\eqref{eq:upper_bound_supersmooth_bias} and~\eqref{eq:upper_bound_supersmooth_variance}, we obtain that
\begin{align*}
    \text{MISE}(p_{N,R}^{\phi}) \leq C_{4} \left(\radius^{\max \{1 - \alpha, 0\}} \exp(-C_{1} \radius^{\alpha}) + \frac{R^{D}}{N} \right).
\end{align*}
As a consequence, we obtain the conclusion of Theorem~\ref{theorem:density_estimation_second} under the supersmooth setting of the function $p$ and $\phi(z) = z$.

\textbf{Ordinary smooth setting of the function $p$:} The proof of Theorem~\ref{theorem:density_estimation_second} when the function $p$ is ordinary smooth also proceeds in the similar fashion as that when $p$ is supersmooth. In particular, we have
\begin{align}
    \int_{\mathbb{R}^{D}} \widehat{p}^2(\bt)(1 - \bar{K}_{R}(\bt))^2 d\bt 
    \leq c \sum_{i = 1}^{D} \int_{B_{i}} \prod_{j = 1}^{D} \frac{1}{(1 + |t_{j}|^{\beta})} d\bt, \label{eq:key_supersmooth_first}
\end{align}
where $B_{i} : = \{\bt \in \mathbb{R}^{D}: \ |t_{i}| \geq R\}$. By simple algebra, we obtain that
\begin{align*}
    \int_{B_{i}} \prod_{j = 1}^{D} \frac{1}{(1 + |t_{j}|^{\beta})} d\bt & = \parenth{ \int_{\mathbb{R}} \frac{1}{1 + |x|^{\beta}} d x}^{D - 1} \cdot \int_{|x| \geq \radius} \frac{1}{1 + |x|^{\beta}} d x \\
    & \leq \parenth{ \int_{\mathbb{R}} \frac{1}{1 + |x|^{\beta}} d x}^{D - 1} \frac{2}{\beta - 1} R^{-\beta + 1}.
\end{align*}
Putting the above results together leads to
\begin{align}
    \int_{\mathbb{R}^{D}} \widehat{p}^2(\bt)(1 - \bar{K}_{R}(\bt))^2 d\bt \leq c_{1} R^{-\beta + 1}, \label{eq:key_ordinary_first}
\end{align}
where $c_{1}$ is some universal constant.

Similar to the supersmooth setting, we also can bound the variance $\frac{1}{N} \int_{\mathbb{R}^{D}} (1 - |\widehat{p}(\bt)|^2) \bar{K}_{R}^2(\bt) d\bt$ under the ordinary smooth setting as follows:
\begin{align}
    \frac{1}{N} \int_{\mathbb{R}^{D}} (1 - |\widehat{p}(\bt)|^2) \bar{K}_{R}^2(\bt) d\bt \leq \frac{R^{D}}{N}. \label{eq:key_ordinary_second}
\end{align}
Combining the results from equations~\eqref{eq:key_ordinary_first} and~\eqref{eq:key_equation_expectation_second}, we obtain that
\begin{align*}
    \text{MISE}(p_{N,R}^{\phi}) \leq c_{2} \left(\radius^{-\beta + 1} + \frac{R^{D}}{N} \right),
\end{align*}
where $c_{2}$ is a universal constant. As a consequence, we obtain the conclusion of Theorem~\ref{theorem:density_estimation_second} under the ordinary smooth setting of the function $p$ and $\phi(z) = z$.
\subsubsection{When $\phi(z) = z^2$}
When $\phi(z) = z^2$, which corresponds to the F\'{e}jer integral setting, we find that
\begin{align*}
    \bar{K}_{R}(t)  = \frac{1}{2^{D}} \prod_{i = 1}^{d} \left(2 - \left|\frac{t_{i}}{R}\right|\right) \bold{1}_{\{|t_{i}| \leq 2R\}}.
\end{align*}
Given the formulation of the function $\bar{K}_{R}$, we first bound $\frac{1}{N} \int_{\mathbb{R}^{D}} (1 - |\widehat{p}(\bt)|^2) \bar{K}_{R}^2(\bt) d\bt$. Indeed, direct calculation shows that
\begin{align}
    \frac{1}{N} \int_{\mathbb{R}^{D}} (1 - |\widehat{p}(\bt)|^2) \bar{K}_{R}^2(\bt) d\bt \leq \frac{1}{N} \int_{\mathbb{R}^{D}} \bar{K}_{R}^2(\bt) d\bt & = \frac{1}{N 2^{D}} \parenth{\int_{|x| \leq 2R} \parenth{2 - \frac{|x|}{R}} d x}^{D} \nonumber \\
    & = \frac{2^{D} R^{D}}{N}. \label{eq:key_Fejer_first}
\end{align}
Now, we proceed to upper bound $\int_{\mathbb{R}^{D}} \widehat{p}^2(\bt)(1 - \bar{K}_{R}(\bt))^2 d\bt$. We have two settings of the function $p$.

\textbf{Supersmooth setting of the function $p$:}
Given the above formulation of the function $\bar{K}_{R}$, we have
\begin{align}
    \int_{\mathbb{R}^{D}} \widehat{p}^2(\bt)(1 - \bar{K}_{R}(\bt))^2 d\bt & = \int_{\mathbb{R}^{D} \backslash [-2R,2R]^{D}} \widehat{p}^2(\bt) d\bt \nonumber \\
    & + \int_{[-2R,2R]^{D}} \widehat{p}^2(\bt) \parenth{1 - \prod_{i = 1}^{D} \parenth{1 - \frac{|t_{i}|}{2R}}}^{2} d\bt. \label{eq:key_supersmooth_Fejer_first}
\end{align}
By using the similar argument as when $\phi(x) =x$, when $p$ is supersmooth function, we obtain that
\begin{align}
\int_{\mathbb{R}^{D} \backslash [-2R,2R]^{D}} \widehat{p}^2(\bt) d\bt \leq C_{1}' \radius^{\max \{1 - \alpha, 0\}} \exp(-C_{2}' \radius^{\alpha}), \label{eq:key_supersmooth_Fejer_second}
\end{align}
where $C_{1}'$ and $C_{2}'$ are universal constants. On the other hand, we have
\begin{align}
    \int_{[-2R,2R]^{D}} \widehat{p}^2(\bt) \parenth{1 - \prod_{i = 1}^{D} \parenth{1 - \frac{|t_{i}|}{2R}}}^{2} d\bt & \nonumber \\
    & \hspace{- 12 em} \leq C_{1} \int_{[-2R,2R]^{D}} \exp \parenth{ -C_{2} \parenth{ \sum_{j = 1}^{D} |t_{j}|^{\alpha}} } \parenth{1 - \prod_{i = 1}^{D} \parenth{1 - \frac{|t_{i}|}{2R}}}^{2} d\bt \nonumber \\
    & \hspace{- 12 em} \leq \bar{C}_{1} \sum_{m = 1}^{D} \sum_{i_{1}, \ldots, i_{m}} \int_{[-2R,2R]^{D}} \exp \parenth{ -C_{2} \parenth{ \sum_{j = 1}^{D} |t_{j}|^{\alpha}} } \frac{\prod_{l = 1}^{m} t_{i_{l}}^2}{R^{2m}} d\bt, \label{eq:key_supersmooth_Fejer_third}
\end{align}
where $\bar{C}_{1}$ is some universal constant. Here, $i_{1}, \ldots, i_{m}$ in the sum satisfy that they are pairwise different and $1 \leq i_{1}, \ldots, i_{m} \leq D$. Now, simple calculations indicate that
\begin{align}
    \int_{[-2R,2R]^{D}} \exp \parenth{ -C_{2} \parenth{ \sum_{j = 1}^{D} |t_{j}|^{\alpha}} } \frac{\prod_{l = 1}^{m} t_{i_{l}}^2}{R^{2m}} d\bt \leq & \nonumber \\
    & \hspace{- 8 em} \frac{1}{R^{2m}} \int_{\mathbb{R}^{D}} \exp \parenth{ -C_{2} \parenth{ \sum_{j = 1}^{D} |t_{j}|^{\alpha}} } \prod_{l = 1}^{m} t_{i_{l}}^2 d \bt \leq \frac{\bar{C}_{2}}{R^{2m}}, \label{eq:key_supersmooth_Fejer_fourth}
\end{align}
where $\bar{C}_{2}$ is some universal constant. Combining the results from equations~\eqref{eq:key_supersmooth_Fejer_third} and~\eqref{eq:key_supersmooth_Fejer_fourth}, there exists universal constant $\bar{C}_{3}$ depending on $D$ such that
\begin{align}
    \int_{[-2R,2R]^{D}} \widehat{p}^2(\bt) \parenth{1 - \prod_{i = 1}^{D} \parenth{1 - \frac{|t_{i}|}{2R}}}^{2} d\bt \leq \frac{\bar{C}_{3}}{R^{2}}. \label{eq:key_supersmooth_Fejer_fifth}
\end{align}
Plugging the inequalities~\eqref{eq:key_supersmooth_Fejer_second} and~\eqref{eq:key_supersmooth_Fejer_fifth} to equation~\eqref{eq:key_supersmooth_Fejer_first} leads to the following bound
\begin{align}
    \int_{\mathbb{R}^{D}} \widehat{p}^2(\bt)(1 - \bar{K}_{R}(\bt))^2 d\bt \leq C_{1}' \radius^{\max \{1 - \alpha, 0\}} \exp(-C_{2}' \radius^{\alpha}) + \frac{\bar{C}_{3}}{R^{2}} \leq \frac{\bar{C}_{4}}{R^2}. \label{eq:key_supersmooth_Fejer_sixth}
\end{align}
Combining the results from equations~\eqref{eq:key_Fejer_first} and~\eqref{eq:key_supersmooth_Fejer_sixth}, we have
\begin{align*}
    \text{MISE}(p_{N,R}^{\phi}) \leq \bar{C}_{5} \parenth{\frac{1}{R^2} + \frac{R^{D}}{N}}.
\end{align*}
As a consequence, we obtain the conclusion of Theorem~\ref{theorem:density_estimation_second} when $\phi(z) = z^2$ and the function $p$ is supersmooth function.

\textbf{Ordinary smooth setting of the function $p$:} Using similar proof argument as that of the supersmooth setting of the function $p$, as $\beta > 3$, we find that
\begin{align}
    \int_{\mathbb{R}^{D}} \widehat{p}^2(\bt)(1 - \bar{K}_{R}(\bt))^2 d\bt & \leq \frac{c}{R^{\beta - 1}} + \int_{[-2R,2R]^{D}} \widehat{p}^2(\bt) \parenth{1 - \prod_{i = 1}^{D} \parenth{1 - \frac{|t_{i}|}{2R}}}^{2} d\bt \nonumber \\
    & \leq \frac{c}{R^{\beta - 1}} + \frac{c_{1}}{R^2} \leq \frac{c_{2}}{R^{2}}, \label{eq:key_ordinary_Fejer_second}
\end{align}
where $c, c_{1}, c_{2}$ are universal constants. Combining the inequalities~\eqref{eq:key_Fejer_first} and~\eqref{eq:key_ordinary_Fejer_second}, we obtain the conclusion of Theorem~\ref{theorem:density_estimation_second} under the ordinary smooth setting of the function $p$ and $\phi(z) = z^2$. 
%%%%%%%%%%%%%%%%%%%%%%%%%%%%%%%%%%%%%%%%%%%%%%%%%%%%%
\subsection{Proof of Theorem~\ref{theorem:nonparametric_regression_first}}
\label{sec:proof:theorem:nonparametric_regression_first}
Our proof strategy is to first bound the bias of $f_{N,R}(\bk)$ and then establish an upper bound for the variance of $f_{N,R}(\bk)$ for each $\bk \in \mathbb{R}^{D}$. 
\subsubsection{Upper bound on the bias} Recall that in equation~\eqref{eq:nonparametric_regression_Fourier}, we define $f_{N,R}(\bk)$ as follows:
\begin{align*}
    f_{N,R}(\bk) : = \frac{\sum_{i = 1}^{N} \bv_{i} \prod_{j = 1}^{D} \phi \left(\frac{\sin(R(k_{j} - k_{ij}))}{R(k_{j} - k_{ij})} \right)}{\sum_{i = 1}^{N} \prod_{j = 1}^{D} \phi \left(\frac{\sin(R(k_{j} - k_{ij}))}{R(k_{j} - k_{ij})} \right)} = \frac{a_{N, R}(\bk)}{p_{N, R}^{\phi}(\bk)},
\end{align*}
where $p_{N, R}^{\phi}(\bk)$ is generalized Fourier density estimator in equation~\eqref{eq:generalized_Fourier_density_estimator} while $a_{N, R}(\bk)$ is defined as follows:
\begin{align*}
    a_{N, R}(\bk) : = \frac{R^{D}}{n A^{D}}\sum_{i = 1}^{N} \bv_{i} \prod_{j = 1}^{D} \phi \left(\frac{\sin(R(k_{j} - k_{ij}))}{R(k_{j} - k_{ij})} \right).
\end{align*}
Simple algebra leads to
\begin{align}
    f_{N,R}(\bk) - f(\bk) = \frac{a_{N,R}(\bk) - f(\bk) p_{N,R}^{\phi}(\bk)}{p(\bk)} + \frac{(f_{N,R}(\bk) - f(\bk)) (p(\bk) - p_{n,R}^{\phi}(\bk))}{p(\bk)}. \label{eq:nonparametric_regression_equation}
\end{align}
Therefore, via an application of Cauchy-Schwarz inequality we obtain that
\begin{align}
    & \hspace{- 2 em} \parenth{ \Exs \brackets{f_{N,R}(\bk)} - f(\bk)}^2 \nonumber \\
    & \leq 2 \frac{\parenth{\Exs \brackets{a_{N,R}(\bk) - f(\bk) p_{N,R}^{\phi}(\bk)}}^2}{p^2(\bk)} + 2 \frac{\parenth{\Exs \brackets{(f_{N,R}(\bk) - f(\bk)) (p(\bk) - p_{N,R}^{\phi}(\bk))}}^2}{p^2(\bk)} \nonumber \\
    & \leq 2 \frac{\parenth{\Exs \brackets{a_{N,R}(\bk) - f(\bk) p_{N,R}^{\phi}(\bk)}}^2}{p^2(\bk)} + 2 \frac{\Exs \brackets{(f_{N, R}(\bk) - f(\bk))^2} \Exs \brackets{(p(\bk) - p_{N,R}^{\phi}(\bk))^2}}{p^2(\bk)}, \label{eq:nonparametric_regression_equation_first}
\end{align}
where the second inequality is due to the standard inequality $\Exs^2(XY) \leq \Exs(X^2)\Exs(Y^2)$ for all the random variables $X, Y$. 

According to the assumptions of Theorem~\ref{theorem:nonparametric_regression_first} and the result of Theorem~\ref{theorem:density_estimation_first}, we have
\begin{align}
    \Exs \brackets{(p(\bk) - p_{N,R}^{\phi}(\bk))^2} \leq \frac{C_{1}}{R^{2(m + 1)}} + \frac{C_{2} R^{D}}{N}, \label{eq:nonparametric_regression_equation_first_extra}
\end{align}
where $C_{1}$ and $C_{2}$ are some universal constants in Theorem~\ref{theorem:density_estimation_first}. 

Now, we proceed to bound $\abss{\Exs \brackets{a_{N,R}(\bk) - f(\bk) p_{N,R}(\bk)}}$. Direct calculation demonstrates that
\begin{align}
    \Exs \brackets{a_{N,R}(\bk)} & = \frac{R^{D}}{A^{D}} \int_{\mathbb{R}^{D}} \prod_{j = 1}^{D} \phi \left(\frac{\sin(R(k_{j} - y_{j}))}{R(k_{j} - y_{j})} \right) p(\by) f(\by) d\by \nonumber \\
    & = \frac{1}{A^{D}} \int_{\mathbb{R}^{D}} \prod_{j = 1}^{D} \phi \left(\frac{\sin(y_{j})}{y_{j}} \right) p \left(\bk - \frac{\by}{R} \right) f\left(\bk - \frac{\by}{R} \right) d\by. \label{eq:nonparametric_regression_equation_second}
\end{align}
An application of Taylor expansion up to the $m$-th order indicates that
\begin{align}
    p \left(\bk - \frac{\by}{R} \right) & = \sum_{0 \leq |\alpha| \leq m} \frac{1}{R^{|\alpha|} \alpha!} \prod_{j = 1}^{D} (-y_{j})^{\alpha_{j}} \dfrac{\partial^{|\alpha|} p}{\partial{\bk^{\alpha}}} (\bk) + \bar{R}_{1}(\bk, \by), \nonumber \\
    f\left(\bk - \frac{\by}{R} \right) & = \sum_{0 \leq |\alpha| \leq m} \frac{1}{R^{|\alpha|} \alpha!} \prod_{j = 1}^{D} (-y_{j})^{\alpha_{j}} \dfrac{\partial^{|\alpha|} f}{\partial{\bk^{\alpha}}} (\bk) + \bar{R}_{2}(\bk, \by) ,\label{eq:nonparametric_regression_equation_third}
\end{align}
where $\alpha = (\alpha_{1}, \ldots, \alpha_{d})$, $|\alpha| = \sum_{j = 1}^{d} \alpha_{j}$, and $\bar{R}_{1}(\bk, \by)$, $R_{2}(\bk, \by)$ are Taylor remainders admitting the following forms:
\begin{align}
    \bar{R}_{1}(\bk, \by) = \sum_{|\beta| = m + 1} \frac{m + 1}{R^{m + 1}\beta!} \prod_{j = 1}^{D} (-y_{j})^{\beta_{j}} \int_{0}^{1} (1 - t)^{m} \dfrac{\partial^{m + 1} p}{\partial{\bk^{\beta}}} \left(\bk - \frac{t \by}{R} \right) dt, \nonumber \\
    \bar{R}_{2}(\bk, \by) = \sum_{|\beta| = m + 1} \frac{m + 1}{R^{m + 1}\beta!} \prod_{j = 1}^{D} (-y_{j})^{\beta_{j}} \int_{0}^{1} (1 - t)^{m} \dfrac{\partial^{m + 1} f}{\partial{\bk^{\beta}}} \left(x - \frac{t \by}{R} \right) dt. \label{eq:nonparametric_regression_equation_fourth}
\end{align}
Combining equations~\eqref{eq:nonparametric_regression_equation_third} and~\eqref{eq:nonparametric_regression_equation_fourth}, we obtain that
\begin{align*}
    p \left(\bk - \frac{\by}{R} \right) f\left(\bk - \frac{\by}{R} \right) & = \sum_{0 \leq |\alpha|, |\beta| \leq m} \frac{1}{R^{|\alpha| + |\beta|} \alpha! \beta!} \prod_{j = 1}^{D} (-y_{j})^{\alpha_{j} + \beta_{j}} \dfrac{\partial^{|\alpha|} p}{\partial{\bk^{\alpha}}} (\bk) \dfrac{\partial^{|\beta|} f}{\partial{\bk^{\beta}}} (\bk) \\
    & + \parenth{\sum_{0 \leq |\alpha| \leq m} \frac{1}{R^{|\alpha|} \alpha!} \prod_{j = 1}^{D} (-y_{j})^{\alpha_{j}} \dfrac{\partial^{|\alpha|} p}{\partial{\bk^{\alpha}}} (\bk)} \bar{R}_{2}(\bk, \by) \\
    & + \parenth{ \sum_{0 \leq |\alpha| \leq m} \frac{1}{R^{|\alpha|} \alpha!} \prod_{j = 1}^{D} (-y_{j})^{\alpha_{j}} \dfrac{\partial^{|\alpha|} f}{\partial{\bk^{\alpha}}} (\bk)} \bar{R}_{1}(\bk, \by) + \bar{R}_{1}(\bk, \by) \bar{R}_{2}(\bk, \by).
\end{align*}
As we have $\int_{\mathbb{R}} \phi \left(\frac{\sin(z)}{z}\right) z^{j} dz = 0$ for all $1 \leq j \leq m$, plugging the equation in the above display to equation~\eqref{eq:nonparametric_regression_equation_second} leads to
\begin{align*}
    \Exs \brackets{a_{n,R}(\bk)} = f(\bk) \Exs \brackets{p_{N, R}^{\phi}(\bk)} + B_{1} + B_{2} + B_{3} + B_{4},
\end{align*}
where $B_{1}, B_{2}, B_{3}, B_{4}$ are defined as follows:
\begin{align*}
    B_{1} & = \frac{1}{A^{D}} \sum_{m + 1 \leq |\alpha| + |\beta| \leq 2m} \int_{\mathbb{R}^{D}} \prod_{j = 1}^{D} \phi \left(\frac{\sin(y_{j})}{y_{j}} \right) \frac{1}{R^{|\alpha| + |\beta|} \alpha! \beta!} \prod_{j = 1}^{D} (-y_{j})^{\alpha_{j} + \beta_{j}} \dfrac{\partial^{|\alpha|} p}{\partial{\bk^{\alpha}}} (\bk) \dfrac{\partial^{|\beta|} f}{\partial{\bk^{\beta}}} (\bk) d\by, \\
    B_{2} & = \frac{1}{A^{D}} \int_{\mathbb{R}^{D}} \prod_{j = 1}^{D} \phi \left(\frac{\sin(y_{j})}{y_{j}} \right) \parenth{\sum_{0 \leq |\alpha| \leq m} \frac{1}{R^{|\alpha|} \alpha!} \prod_{j = 1}^{D} (-y_{j})^{\alpha_{j}} \dfrac{\partial^{|\alpha|} p_{0}}{\partial{\bk^{\alpha}}} (\bk)} \bar{R}_{2}(\bk, \by) d\by, \\
    B_{3} & = \frac{1}{A^{D}} \int_{\mathbb{R}^{D}} \prod_{j = 1}^{D} \phi \left(\frac{\sin(y_{j})}{y_{j}} \right) \parenth{\sum_{0 \leq |\alpha| \leq m} \frac{1}{R^{|\alpha|} \alpha!} \prod_{j = 1}^{D} (-y_{j})^{\alpha_{j}} \dfrac{\partial^{|\alpha|} f}{\partial{\bk^{\alpha}}} (\bk)} \bar{R}_{1}(\bk, \by) d\by, \\
    B_{4} & = \frac{1}{A^{D}} \int_{\mathbb{R}^{D}} \prod_{j = 1}^{D} \phi \left(\frac{\sin(y_{j})}{y_{j}} \right) \bar{R}_{1}(\bk, \by) \bar{R}_{2}(\bk, \by) d\by.
\end{align*}
Since we have $\int_{\mathbb{R}} \left|\phi \left(\frac{\sin(z)}{z}\right)\right| |z|^{j} dz < \infty$ for any $m + 1 \leq j \leq 2m + 2$ and $p_{0}, f \in \mathcal{C}^{m+1}(\mathbb{R}^{d})$, we find that as long as $R \geq \bar{c}$ for some given constant $\bar{c}$
\begin{align*}
    |B_{1}| & \leq \frac{1}{A^{D}} \sum_{m + 1 \leq |\alpha| + |\beta| \leq 2m} \frac{1}{R^{|\alpha| + |\beta|} \alpha! \beta!} \int_{\mathbb{R}^{D}} \prod_{j = 1}^{D} \abss{\phi \left(\frac{\sin(y_{j})}{y_{j}} \right)} \prod_{j = 1}^{D} |y_{j}|^{\alpha_{j} + \beta_{j}} \|\dfrac{\partial^{|\alpha|} p}{\partial{\bk^{\alpha}}}\|_{\infty} \|\dfrac{\partial^{|\beta|} f}{\partial{\bk^{\beta}}}\|_{\infty} \\
    & \leq \frac{c_{1}}{R^{m + 1}},
\end{align*}
where $c_{1}$ is some universal constant depending on $A$, $D$, and $\bar{c}$. Furthermore, we find that
\begin{align*}
    |B_{2}| & \leq \frac{1}{A^{D}} \sum_{0 \leq |\alpha| \leq m, |\beta| = m + 1} \frac{m + 1}{R^{|\alpha| + m + 1} \alpha! \beta!} \int_{\mathbb{R}^{D}} \prod_{j = 1}^{D} \abss{\phi \left(\frac{\sin(y_{j})}{y_{j}} \right)} \prod_{j = 1}^{D} |y_{j}|^{\alpha_{j} + \beta_{j}} \\
    & \hspace{10 em} \times \int_{0}^{1} (1 - t)^{m} \|\dfrac{\partial^{m + 1} f}{\partial{\bk^{\beta}}}\|_{\infty} d\by dt \leq \frac{c_{2}}{R^{m + 1}},
\end{align*}
where $c_{2}$ is some universal constant depending on $A$, $d$, and $\bar{c}$. Similarly, we also can demonstrate that $B_{3} \leq c_{3}/ R^{m + 1}$ and $B_{4} \leq c_{4}/ R^{2(m + 1)}$ for some universal constants $c_{3}$ and $c_{4}$. Putting the above results together, we arrive at the following bound:
\begin{align}
    \abss{\Exs \brackets{a_{n,R}(\bk) - f(\bk) p_{N,R}^{\phi}(\bk)}} \leq \frac{c'}{R^{m + 1}}. \label{eq:nonparametric_regression_equation_fifth}
\end{align}
Plugging the results from equations~\eqref{eq:nonparametric_regression_equation_first_extra} and~\eqref{eq:nonparametric_regression_equation_fifth} to equation~\eqref{eq:nonparametric_regression_equation_first}, we obtain that
\begin{align}
    \parenth{ \Exs \brackets{f_{N,R}(\bk)} - f(\bk)}^2 \leq \frac{2 (c')^2}{p^2(\bk) R^{2(m+1)}} + \frac{2 \Exs \brackets{(f_{N, R}(\bk) - f(\bk))^2}}{p^2(\bk)} \parenth{\frac{C_{1}}{R^{2(m + 1)}} + \frac{C_{2} R^{D}}{N}}. \label{eq:bound_bias_nonparametric}
\end{align}
\subsubsection{Upper bound on the variance} Now, we study the variance of $f_{N,R}(\bk)$.  By taking variance both sides of the equation~\eqref{eq:nonparametric_regression_equation}, we obtain that
\begin{align}
    \var(f_{N,R}(\bk)) & = \var \parenth{\frac{a_{N,R}(\bk) - f(\bk) p_{N,R}^{\phi}(\bk)}{p(\bk)} + \frac{(f_{N,R}(\bk) - f(\bk)) (p(\bk) - p_{N,R}^{\phi}(\bk))}{p(\bk)}} \nonumber \\
    & \hspace{- 4 em} \leq \frac{2}{p^2(\bk)} \parenth{ \underbrace{\Exs \brackets{ \parenth{a_{N,R}(\bk) - f(\bk) p_{N,R}^{\phi}(\bk)}^2}}_{T_{1}} + \underbrace{\Exs \brackets{(f_{N,R}(\bk) - f(\bk))^2 (p(\bk) - p_{N,\radius}^{\phi}(\bk))^2}}_{T_{2}}}. \label{eq:bound_var_nonparametric_regression}
\end{align}
\textbf{Upper bound of $T_{2}$:} To upper bound $T_{2}$, we utilize the following lemma.
\begin{lemma}
\label{lemma:concentration_Fourier_estimator}
Assume that the function $\phi$ and $p_{0}$ satisfy the assumptions of Theorem~\ref{theorem:density_estimation_first}. Furthermore, $\phi(z) \leq C$ as long as $|z| \leq 1$ for some universal constant $C$. Then, for almost all $\bk \in \mathbb{R}^{D}$, there exist universal constants $C'$ such that
\begin{align*}
    \Prob \parenth{\abss{ p_{N, \radius}^{\phi}(\bk) - p(\bk)} \geq C' \parenth{\frac{1}{R^{m + 1}} + \sqrt{\frac{\radius^{D} \log(2/ \delta)}{N}} }} \leq \delta. 
\end{align*}
\end{lemma}
Proof of Lemma~\ref{lemma:concentration_Fourier_estimator} is given in Appendix~\ref{subsec:proof:lemma:concentration_Fourier_estimator}. Now given the result of Lemma~\ref{lemma:concentration_Fourier_estimator}, we denote $B$ as the event such that $$\abss{p_{N, \radius}^{\phi}(\bk) - p(\bk)} \leq C' \parenth{\frac{1}{R^{m + 1}} + \sqrt{\frac{\radius^{D} \log(2/ \delta)}{N}}}$$ where $C'$ is a universal constant in Lemma~\ref{lemma:concentration_Fourier_estimator}. Then, we obtain $\Prob(B) \geq 1 - \delta$. Hence, we have the following bound with $T_{2}$:
\begin{align*}
    T_{2} & = \Exs \brackets{(f_{N,R}(\bk) - f(\bk))^2 (p(\bk) - p_{N,\radius}^{\phi}(\bk))^2|B} \Prob(B) \\
    & \hspace{ 16 em} + \Exs \brackets{(f_{N,R}(\bk) - f(\bk))^2 (p(\bk) - p_{N,\radius}^{\phi}(\bk))^2| B^{c}} \Prob(B^{c}) \\
    & \leq 2 c' \Exs \brackets{(f_{N,R}(\bk) - f(\bk))^2} \parenth{ \frac{1}{R^{2(m + 1)}} + \frac{\radius^{D} \log(2/ \delta)}{N} + \delta \parenth{p^2(\bk) + \frac{C^{D}\radius^{2D}}{A^{D}}}},
\end{align*}
where $c'$ is some universal constant and the final inequality is based on the inequalities: $\Prob(B^{c}) \leq \delta$ and $(p(\bk) - p_{N,R}^{\phi}(\bk))^2 \leq 2 ( p^2(\bk) + (p_{N,R}^{\phi})^2(\bk)) \leq 2 \parenth{p^2(\bk) + \frac{C^{D} \radius^{2D}}{A^{D}}}$ where $C$ is a universal constant such that $\phi(z) \leq C$ when $|z| \leq 1$. By choosing $\delta$ such that $\delta = \frac{\radius^{D}}{N(p^2(\bk) + C^{D} \radius^{2D}/ A^{D})}$, we obtain that
\begin{align}
    T_{2} \leq c'' \Exs \brackets{(f_{N,R}(\bk) - f(\bk))^2} \parenth{ \frac{1}{R^{2(m + 1)}} + \frac{\radius^{D} \log(N R)}{N}}, \label{eq:bound_T2}
\end{align}
for some universal constant $c''$ when $\radius$ is sufficiently large.

\textbf{Upper bound of $T_{1}$:} As $\bv_{i} = f(\bk_{i}) + \epsilon_{i}$ for all $i \in [N]$, direct calculation shows that
\begin{align*}
    T_{1} & = \mathbb{E} \biggr[ \biggr(\frac{R^{D}}{N A^{D}} \sum_{i = 1}^{N} (f(\bk_{i}) - f(\bk)) \prod_{j = 1}^{D} \phi \left(\frac{\sin(R(k_{j} - k_{ij}))}{R(k_{j} - k_{ij})} \right) \\ 
    & \hspace{12 em} + \frac{R^{D}}{N A^{D}} \sum_{i = 1}^{N} \epsilon_{i} \prod_{j = 1}^{D} \phi \left(\frac{\sin(R(k_{j} - k_{ij}))}{R(k_{j} - k_{ij})} \right) \biggr)^2 \biggr].
\end{align*}
An application of Cauchy-Schwarz inequality leads to
\begin{align*}
    T_{1} & \leq 2 \Exs \brackets{\parenth{\frac{R^{D}}{N A^{D}} \sum_{i = 1}^{N} (f(\bk_{i}) - f(\bk)) \prod_{j = 1}^{D} \phi \left(\frac{\sin(R(k_{j} - k_{ij}))}{R(k_{j} - k_{ij})} \right)}^2 } \\
    & \hspace{ 8 em} + 2 \Exs \brackets{ \parenth{\frac{1}{N \pi^{D}} \sum_{i = 1}^{N} \epsilon_{i} \prod_{j = 1}^{D} \phi \left(\frac{\sin(R(k_{j} - k_{ij}))}{R(k_{j} - k_{ij})} \right)}^2 } = 2( S_{1} + S_{2}).
\end{align*}
Since we have $\Exs \brackets{ \parenth{\frac{1}{N} \sum_{i = 1}^{N} Z_{i} }^2} \leq \frac{1}{N} \Exs \brackets{Z_{1}^2} + \Exs^2 \brackets{Z_{1}}$ for any i.i.d. samples $Z_{1},\ldots, Z_{N}$, we obtain that
\begin{align*}
    S_{1} & \leq \frac{R^{2D}}{N A^{2D}} \Exs \brackets{(f(X) - f(\bk))^2 \prod_{j = 1}^{D} \phi^2 \left(\frac{\sin(R(k_{j} - X_{.j}))}{R(k_{j} - X_{.j})}\right)} & \\
    & \hspace{8 em} + \frac{R^{2D}}{A^{2D}} \Exs^2 \brackets{(f(X) - f(\bk)) \prod_{j = 1}^{D} \phi \left(\frac{\sin(R(k_{j} - X_{.j}))}{R(k_{j} - X_{.j})}\right)},
\end{align*}
where the outer expectation is taken with respect to $X = (X_{.1}, \ldots, X_{.d}) \sim p$. From the result in equation~\eqref{eq:nonparametric_regression_equation_fifth}, we have 
\begin{align*}
    \frac{R^{2D}}{A^{2D}} \Exs^2 \brackets{(f(X) - f(\bk)) \prod_{j = 1}^{D} \phi \left(\frac{\sin(R(k_{j} - X_{.j}))}{R(k_{j} - X_{.j})}\right)} & = \Exs^2 \brackets{a_{N,R}(\bk) - f(\bk) p_{N,R}^{\phi}(\bk)} \leq \frac{c'}{R^{2(m + 1)}},
\end{align*}
where $c'$ is some universal constant. In addition, an application of Cauchy-Schwarz inequality leads to
\begin{align*}
    & \hspace{- 5 em} \frac{R^{2D}}{N A^{2D}} \Exs \brackets{(f(X) - f(\bk))^2 \prod_{j = 1}^{D} \phi^2 \left(\frac{\sin(R(k_{j} - X_{.j}))}{R(k_{j} - X_{.j})}\right)} \\
    & \leq \frac{2 R^{2D}}{N A^{2D}} \Exs \brackets{(f^{2}(X) + f^{2}(\bk)) \prod_{j = 1}^{D} \phi^2 \left(\frac{\sin(R(k_{j} - X_{.j}))}{R(k_{j} - X_{.j})}\right)} \\
    & = \frac{2 R^{D}}{N A^{2D}} \int_{\mathbb{R}^{D}} \prod_{j = 1}^{D} \phi^2 \left(\frac{\sin(y_{j}))}{y_{j}}\right) \biggr(f^2 \left(\bk - \frac{\by}{R} \right) p \left(\bk - \frac{\by}{R} \right) + f^2(\bk) \biggr) d\by \\
    & \leq \frac{2 R^{D}(\|f^2 \times p\|_{\infty} + f^2(\bk))}{N A^{2D}} \int_{\mathbb{R}^{D}} \prod_{j = 1}^{D} \phi^2 \left(\frac{\sin(y_{j}))}{y_{j}}\right) dy.
\end{align*}
Since we have $\int_{\mathbb{R}} \phi^2(\sin(z)/z) dz < \infty$, it indicates that we can find a universal constant $c''$ such that
\begin{align*}
    \frac{R^{2D}}{N A^{2D}} \Exs \brackets{(f(X) - f(\bk))^2 \prod_{j = 1}^{D} \phi^2 \left(\frac{\sin(R(k_{j} - X_{.j}))}{R(k_{j} - X_{.j})}\right)} \leq \frac{c'' R^{D} (\|f^2 \times p\|_{\infty} + f^2(\bk))}{N A^{2D}}. 
\end{align*}
Putting the above results together, we obtain that
\begin{align}
    S_{1} \leq \frac{c'}{R^{2(m + 1)}} + \frac{c'' R^{D} (\|f^2 \times p\|_{\infty} + f^2(\bk))}{N A^{2D}}. \label{eq:bound_S1}
\end{align}
Similarly, since $\Exs(\epsilon_{i}) = 0$ and $\var(\epsilon_{i}) = \sigma^2$ for all $i \in [N]$, we have
\begin{align}
    S_{2} = \frac{\sigma^2 R^{2D}}{N A^{2D}} \Exs \brackets{\prod_{j = 1}^{D} \phi^2 \left(\frac{\sin(R(k_{j} - X_{.j}))}{R(k_{j} - X_{.j})}\right)} \leq \frac{c''' \sigma^2 R^{D} \|p\|_{\infty} \radius^{D}}{N A^{2D}}, \label{eq:bound_S2}
\end{align}
where $c'''$ is some universal constant. Combining the results from equation~\eqref{eq:bound_S1} and equation~\eqref{eq:bound_S2}, we find that
\begin{align}
    T_{1} \leq C \parenth{\frac{(\|f^2 \times p\|_{\infty} + f^2(\bk) + \sigma^2 \|p\|_{\infty}) R^{D}}{N} + \frac{1}{R^{2(m+1)}}}, \label{eq:bound_T1}
\end{align}
where $C$ is some universal constant. Plugging the bounds of $T_{1}$ and $T_{2}$ from equations~\eqref{eq:bound_T2} and~\eqref{eq:bound_T1} into equation~\eqref{eq:bound_var_nonparametric_regression}, when $\radius \geq C'$ where $C'$ is some universal constant, we have 
\begin{align}
    \var(f_{N,R}(\bk)) \leq \frac{C_{1}'}{p^2(\bk)} \Exs \brackets{(f_{N,R}(\bk) - f(\bk))^2} \parenth{ \frac{1}{R^{2(m + 1)}} + \frac{\radius^{D} \log(N R)}{N}} & \nonumber \\
    & \hspace{- 20 em} + \frac{C_{2}'}{p^2(\bk)} \parenth{\frac{(f(\bk) + C_{3}') \radius^{D}}{N} + \frac{1}{R^{2(m + 1)}}}, \label{eq:var_bound_nonparametric_regression}
\end{align}
where $C_{1}', C_{2}', C_{3}'$ are some universal constants. Combining the results with bias and variance in equations~\eqref{eq:bound_bias_nonparametric} and~\eqref{eq:var_bound_nonparametric_regression}, we obtain the following bound:
\begin{align*}
    \Exs \brackets{(f_{N,R}(\bk) - f(\bk))^2} & \leq \frac{2 (c')^2}{p^2(\bk) R^{2(m+1)}} + \frac{2 \Exs \brackets{(f_{N, R}(\bk) - f(\bk))^2}}{p^2(\bk)} \parenth{\frac{C_{1}}{R^{2(m + 1)}} + \frac{C_{2} R^{D}}{N}} \\
    & + \frac{C_{1}'}{p^2(\bk)} \Exs \brackets{(f_{N,R}(\bk) - f(\bk))^2} \parenth{ \frac{1}{R^{2(m + 1)}} + \frac{\radius^{D} \log(N R)}{N}} \\
    & + \frac{C_{2}'}{p^2(\bk)} \parenth{\frac{(f(\bk) + C_{3}') \radius^{D}}{N} + \frac{1}{R^{2(m + 1)}}}.
\end{align*}
As a consequence, we obtain the conclusion of the theorem.
\subsection{Proof of Lemma~\ref{lemma:concentration_Fourier_estimator}}
\label{subsec:proof:lemma:concentration_Fourier_estimator}
Invoking triangle inequality, we obtain that
\begin{align}
    \abss{ p_{N, \radius}^{\phi}(\bk) - p(\bk)} \leq \abss{p_{N, \radius}^{\phi}(\bk) - \Exs \brackets{p_{N, \radius}^{\phi}(\bk)}} + \abss{\Exs \brackets{p_{N, \radius}^{\phi}(\bk)} - p(\bk)}. \label{eq:concentration_zero}
\end{align}
If we denote $\bv_{i} = \frac{R^{D}}{A^{D}}\prod_{j = 1}^{D} \phi \parenth{\frac{\sin(\radius(k_{j} - k_{ij})}{R(k_{j} - k_{ij})}}$ for all $i \in [N]$, then as $\sin(\radius(k_{j} - k_{ij})/(R(k_{j} - k_{ij})) \leq 1$ for all $j \in [D]$ we have $|\bv_{i}| \leq C^{D} \radius^{D}/ A^{D}$ for all $i \in [N]$ where $C$ is the constant such that $\phi(z) \leq C$ when $|z| \leq 1$. Furthermore, from the proof of Theorem~\ref{theorem:density_estimation_first} we have $\var(\bv_{i}) \leq C' \radius^{D}$ where $C' > 0$ is some universal constant. Given these bounds of $\bv_{i}$ and $\var(\bv_{i})$, for any $t \in (0, C'']$ Bernstein's inequality shows that
\begin{align*}
    \Prob \parenth{\abss{\frac{1}{N} \sum_{i = 1}^{N} \bv_{i} - \Exs \brackets{\bv_{1}}} \geq t} \leq 2 \exp \parenth{- \frac{N t^2}{2 C' \radius^{D} + 2 C^{D} \radius^{D} t/ (3 A^{D})}}.
\end{align*}
By choosing $t = \bar{C} \sqrt{\radius^{D} \log(2/ \delta)/N}$, where $\bar{C}$ is some universal constant, we find that
\begin{align}
    \Prob \parenth{\abss{p_{N, \radius}^{\phi}(\bk) - \Exs \brackets{p_{N, \radius}^{\phi}(\bk)}} \geq t} = \Prob \parenth{\abss{\frac{1}{N} \sum_{i = 1}^{N} \bv_{i} - \Exs \brackets{\bv_{1}}} \geq t} \leq \delta. \label{eq:concentration_first}
\end{align}
From the result of Theorem~\ref{theorem:density_estimation_first}, there exists universal constant $c$ such that
\begin{align}
    \abss{\Exs \brackets{p_{N, \radius}^{\phi}(\bk)} - p(\bk)} \leq c/ R^{m + 1}. \label{eq:concentration_second}
\end{align}
Plugging the bounds~\eqref{eq:concentration_first} and~\eqref{eq:concentration_second} into the triangle inequality~\eqref{eq:concentration_zero}, we obtain the conclusion of the lemma.

\section{Experiment Details}
\label{secapp:expdetails}
\subsection{Language Modeling on WikiText-103}
In our experiments on WikiText-103 in Section~\ref{subsec:wikitext}, we let R be a learnable scalar initialized to 2 and choose $\phi(x) = x^4$. The same setting is used for all attention units in the model; each unit has a different R. We observe that by setting R to be a learnable vector $[R_{1},\dots,R_{D}]^{\top}$, the FourierFormer gains advantage in accuracy but with the cost of the increase in the number of parameters. When R is a vector $[R_{1},\dots,R_{D}]^{\top}$, the equation of the Fourier Attention is given by

\begin{align}
    \hat{\bh}_i := f_{N,R}(\bq_{i}) = \frac{\sum_{i = 1}^{N} \bv_{i} \prod_{j = 1}^{D} \phi \left(\frac{\sin(R_j(q_{ij} - k_{ij}))}{R_j(q_{ij} - k_{ij})} \right)}{\sum_{i = 1}^{N} \prod_{j = 1}^{D} \phi \left(\frac{\sin(R_j(q_{ij} - k_{ij}))}{R_j(q_{ij} - k_{ij})} \right)} \quad \quad \forall \ i \in [N]. \label{eqn:fourier-attention}
\end{align}

We provide an ablation study for the effect of $R$ and $\phi$ in Section~\ref{secapp:more-exp-results} below.

\subsection{Image Classification on ImageNet}
Similar to setting for language modeling, in our experiments on ImageNet image classification, we set R to be a learnable scalar initialized to 1 and choose $\phi(x) = x^4$. Different attention units have different R.

\section{Additional Experimental Results}
\label{secapp:more-exp-results}
\subsection{Effect of $\phi$}
\label{secapp:ablation-phi}
Using the WikiText-103 language modeling as a case study, we analyze the effect of $\phi(x)$ on the performance of FourierFormer. In particular, we set $\phi(x) = x^{k}$ and compare the performance of FourierFormer for $k=1,2,3,4$ and $6$. We keep other settings the same as in our experiments in Section~\ref{subsec:wikitext}. We summarize our results in Table~\ref{tab:lm-results-ablation-phi}. We observe that for odd values of $k$ such as $k=1,3$, the training diverges, confirming that negative density estimator cause instability in training FourierFormer (see Remark~\ref{subsec:generalized-fourier}). For even values of $k$ such as $k=2,4,6$, we observe that the greater value of $k$ results in better valid and test PPL. However, the gap between $k=4$ and $k=6$ is smaller compared to the gap between $k=2$ and $k=4$, suggesting that using $k>4$ does not add much advantage in terms of accuracy.

\begin{table}[t!]
\small
    \caption{\small Ablation study on how the choice of $\phi(x)=x^{k}$ influences the performance of FourierFormer. Odd values of $k$ cause training to diverge. For even values of $k$, greater $k$ yields better perplexity (PPL), but the improvement is small for $k>4$.}
    \vspace{0.1in}
    \begin{center}
    \scalebox{0.9}{\begin{tabular}{lcc}
    \toprule
        Method & Valid PPL & Test PPL \\
        \midrule
        {\it Baseline dot-product (small)} & 33.15  & 34.29 \\
        \midrule
        FourierFormer, $\phi(x) = x^{2}$ (small) & 32.09  & 33.10 \\
        FourierFormer, $\phi(x) = x^{4}$ (small) & 31.86  & 32.85 \\
        FourierFormer, $\phi(x) = x^{6}$ (small) & \bf 31.84  & \bf 32.81 \\
        \midrule
        FourierFormer, $\phi(x) = x$ (small) & not converge  & not converge \\
        FourierFormer, $\phi(x) = x^{3}$ (small) & not converge  & not converge \\
        \bottomrule
    \end{tabular}}
    \end{center}
\label{tab:lm-results-ablation-phi}
\end{table}

\begin{table}[t!]
\small
    \caption{\small Ablation study on how the initialization of $R$ influences the performance of FourierFormer. When $R$ is initialized to a too small or too big value, the PPL of the trained FourierFormer is reduced. $R_{\text{init}} = 1,2,3$ yield the best results. Fourierformer with learnable vectors $R$ yields better results than Fourierformer of the same setting using learnable scalars $R$ with the cost of increasing the number of parameters in the model.}
    \vspace{0.1in}
    \begin{center}
    \scalebox{0.9}{\begin{tabular}{lcc}
    \toprule
        Method & Valid PPL & Test PPL \\
        \midrule
        {\it Baseline dot-product (small)} & 33.15  & 34.29 \\
        \midrule
        FourierFormer, $R_{\text{init}} = 0.1$ (small) & 32.04 & 33.01  \\
        FourierFormer, $R_{\text{init}} = 1.0$ (small) &  31.89 &  32.87 \\
        FourierFormer, $R_{\text{init}} = 2.0$ (small) & \bf 31.86  & \bf 32.85 \\
        FourierFormer, $R_{\text{init}} = 3.0$ (small) & 31.90  & 32.88 \\
        FourierFormer, $R_{\text{init}} = 4.0$ (small) & 32.58  & 33.65 \\
        \midrule
        FourierFormer, $R_{\text{init}} = 2.0$ (small, $R$ is a vector) & \bf 31.82  & \bf 32.80 \\
        \bottomrule
    \end{tabular}}
    \end{center}
\label{tab:lm-results-ablation-R}
\end{table}

\subsection{Effect of the Initialization of $R$}
\label{secapp:ablation-R}
In this section, we study the effect of the initialization value of $R$ on the performance of FourierFormer when trained for the WikiText-103 language modeling and summarize our results in Table~\ref{tab:lm-results-ablation-R}. Here we choose $R$ to be learnable scalars as in experiments described in our main text. Other settings are also the same as in our experiments in Section~\ref{subsec:wikitext}. We observe that when $R$ is initialized too small (e.g. $R_{\text{init}} = 0.1$) or too big (e.g. $R_{\text{init}} = 4$), the PPL of the trained FourierFormer decreases. $R_{\text{init}} = 1,2,3$ yield best results.

We also study the performance of the FourierFormer when R is chosen to be a learnable vector, $R = [R_{1},\dots,R_{D}]^{\top}$. We report our result in the last row of Table~\ref{tab:lm-results-ablation-R}. FourierFormer with R be learnable vectors achieves better PPLs than FourierFormer with R be learnable scalars of the same setting.  As we mentioned in Section~\ref{secapp:expdetails}, this advantage comes with an increase in the number of parameters in the model.

Finally, from our experiments, we observe that making R a learnable parameter yields better PPLs than making R a constant and selecting its value via a careful search.

%%%%%%%%%%%%%%%%%%%%%%%%%%%%%%%%%%%%%%%%%%%%%%%%%%%%%%%%%%%%%%%%%%%%%%%%%
\bibliographystyle{abbrv}
\bibliography{references}
\end{document}

%% file: final_macros.tex
\newcommand{\brackets}[1]{\left[ #1 \right]}
\newcommand{\parenth}[1]{\left( #1 \right)}

\newcommand{\abss}[1]{\left| #1 \right |}

\newcommand{\radius}{\ensuremath{R}}

\newcommand{\Exs}{\ensuremath{{\mathbb{E}}}}
\newcommand{\Prob}{\ensuremath{{\mathbb{P}}}}

\DeclareMathOperator{\var}{var}

\newlength{\widebarargwidth}
\newlength{\widebarargheight}
\newlength{\widebarargdepth}

\makeatletter
\long\def\@makecaption#1#2{
        \vskip 0.8ex
        \setbox\@tempboxa\hbox{\small {\bf #1:} #2}
        \parindent 1.5em  %% How can we use the global value of this???
        \dimen0=\hsize
        \advance\dimen0 by -3em
        \ifdim \wd\@tempboxa >\dimen0
                \hbox to \hsize{
                        \parindent 0em
                        \hfil
                        \parbox{\dimen0}{\def\baselinestretch{0.96}\small
                                {\bf #1.} #2
                                }
                        \hfil}
        \else \hbox to \hsize{\hfil \box\@tempboxa \hfil}
        \fi
        }
\makeatother

\long\def\comment#1{}
\definecolor{battleshipgrey}{rgb}{0.52, 0.52, 0.51}
\definecolor{darkgray}{rgb}{0.66, 0.66, 0.66}
\definecolor{darkgreen}{rgb}{0.0, 0.2, 0.13}
\definecolor{darkspringgreen}{rgb}{0.09, 0.45, 0.27}
\definecolor{dukeblue}{rgb}{0.0, 0.0, 0.61}
\definecolor{olivedrab7}{rgb}{0.24, 0.2, 0.12}
\definecolor{darkblue}{rgb}{0.0, 0.0, 0.55}
\definecolor{darkscarlet}{rgb}{0.34, 0.01, 0.1}
\definecolor{candyapplered}{rgb}{1.0, 0.03, 0.0}
\definecolor{ao(english)}{rgb}{0.0, 0.5, 0.0}
\definecolor{applegreen}{rgb}{0.55, 0.71, 0.0}